\documentclass[10pt,twocolumn,letterpaper]{article}

\usepackage[pagenumbers]{cvpr} 

\usepackage[final]{microtype}
\newif\ifdraft
\drafttrue        

\ifdraft
  \newcommand{\zhengqi}[1]{\textcolor{purple}{\textbf{zhengqi:}~#1}}
  \newcommand{\andy}[1]{\textcolor{teal}{\textbf{andy:}~#1}}
  \newcommand{\ync}[1]{\textcolor{blue}{\textbf{yotam:}~#1}}
  
  \newcommand{\richard}[1]{\textcolor{orange}{\textbf{richard:}~#1}}
  \newcommand{\xjqi}[1]{\textcolor[rgb]{1.0,0,0}{{[\textbf{xjqi}: #1]}}}
  \newcommand{\steve}[1]{\textcolor[rgb]{0.7,0.2,0.1}{{[\textbf{steve}: #1]}}}
\else
  \newcommand{\zhengqi}[1]{}
  \newcommand{\andy}[1]{}
  \newcommand{\ync}[1]{}
  \newcommand{\richard}[1]{}
  \newcommand{\xjqi}[1]{}
  \newcommand{\steve}[1]{}
  
\fi

\newcommand{\NetWork}{V_{\theta}}

\usepackage{soul}

\definecolor{cvprblue}{rgb}{0.21,0.49,0.74}
\usepackage[pagebackref,breaklinks,colorlinks,allcolors=cvprblue]{hyperref}
\renewcommand{\arraystretch}{1.15}
\usepackage{adjustbox}
\usepackage{booktabs}
\usepackage{xcolor,colortbl}
\usepackage{multirow}
\usepackage{adjustbox}
\definecolor{headercolor}{RGB}{220,220,220}
\definecolor{highlightcolor}{RGB}{240,240,240}
\definecolor{rowcolor}{RGB}{245,245,245}

\def\paperID{26} 
\def\confName{CVPR}
\def\confYear{2026}

\usepackage{comment}
\usepackage{amssymb}

\usepackage[dvipsnames]{xcolor}
\usepackage{xfp}
\usepackage{pgffor}
\definecolor{linkpink}{RGB}{220, 80, 120}

\definecolor{placeholdercolor}{rgb}{0.7, 0.7, 0.7}

\usepackage{tikz}
\usepackage{lipsum}

\title{Self-Evaluation Unlocks Any-Step Text-to-Image Generation}

\author{
Xin Yu\textsuperscript{1,2}
\quad
Xiaojuan Qi\textsuperscript{1}%
\thanks{Corresponding author.}%
\;\;\!
\thanks{Project lead.}%
\quad
Zhengqi Li\textsuperscript{2}
\quad
Kai Zhang\textsuperscript{2}
\quad
Richard Zhang\textsuperscript{2}
\\
Zhe Lin\textsuperscript{2}
\quad
Eli Shechtman\textsuperscript{2}
\quad
Tianyu Wang\textsuperscript{2}\footnotemark[2]
\quad
Yotam Nitzan\textsuperscript{2}\footnotemark[2]
\\[4pt]
\textsuperscript{1}The University of Hong Kong
\quad
\textsuperscript{2}Adobe Research
}

\begin{document}
\maketitle
\begin{abstract}

We introduce the Self-Evaluating Model (Self-E), a novel, from-scratch training approach for text-to-image generation that supports any-step inference.
Self-E learns from data similarly to a Flow Matching model, while simultaneously employing a novel self-evaluation mechanism: it evaluates its own generated samples using its current score estimates, effectively serving as a dynamic self-teacher.
Unlike traditional diffusion or flow models, it does not rely solely on local supervision, which typically necessitates many inference steps.
Unlike distillation-based approaches, it does not require a pretrained teacher.
This combination of instantaneous local learning and self-driven global matching bridges the gap between the two paradigms, enabling the training of a high-quality text-to-image model from scratch that excels even at very low step counts.
Extensive experiments on large-scale text-to-image benchmarks show that Self-E not only excels in few-step generation, but is also competitive with state-of-the-art Flow Matching models at 50 steps.
We further find that its performance improves monotonically as inference steps increase, enabling both ultra-fast few-step generation and high-quality long-trajectory sampling within a single unified model.
To our knowledge, Self-E is the first from-scratch, any-step text-to-image model, offering a unified framework for efficient and scalable generation.

\end{abstract}

\section{Introduction}
\label{sec:intro}
\begin{figure*}[t]
  \centering
  \includegraphics[width=\textwidth]{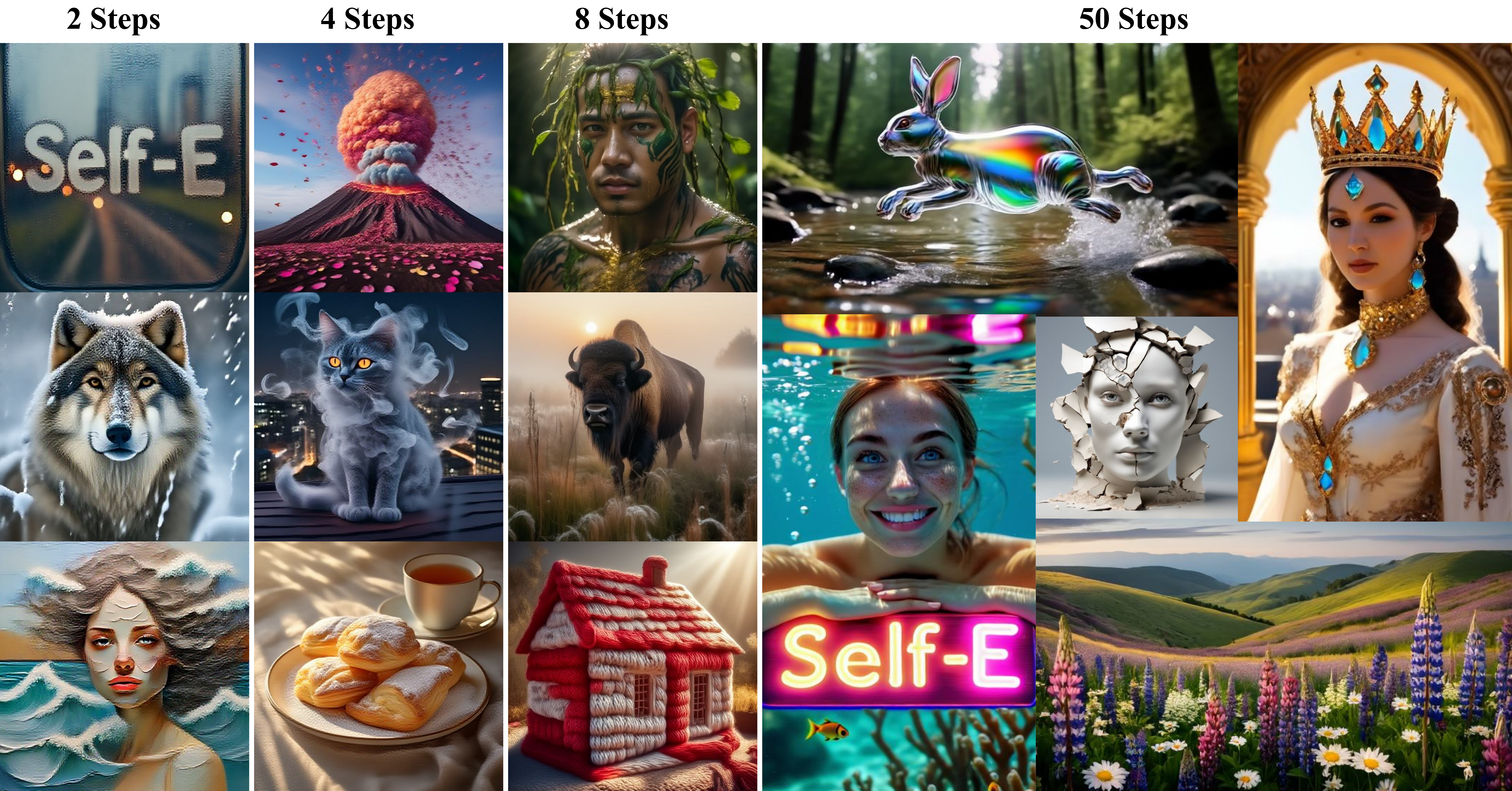}
  \caption{\textbf{Qualitative Any-Step Generation.} We showcase diverse text-to-image results from our model at different inference step counts, demonstrating coherent semantics, strong text alignment. Text prompts are provided in the supplementary material.
}
  \label{fig:teaser}
\end{figure*}

Diffusion models \cite{ho2020denoising,song2020generative,song2020score} and flow matching models \cite{flow_matching,instaflow} currently dominate text-to-image generation due to their stability, scalability, and strong visual fidelity \cite{flux,xie2024sana,xie2025sana,rombach2022high}.
These models are trained to approximate local supervision from data -- either the score function or the instantaneous velocity field -- which specifies how a noisy sample should infinitesimally move toward the data manifold at each timestep.
Because this supervision is inherently local, it provides only short-range guidance: each update corrects small deviations but lacks a holistic global view of the target distribution.
Consequently, diffusion and flow-based models typically require dozens of sequential steps to reliably traverse the curved reverse trajectory from noise to data, making inference computationally expensive and limiting their use in time-sensitive applications.

A dominant strategy for reducing inference steps is distillation, where a pretrained teacher supervises a student model~\cite{salimans2022progressive,meng2023distillation,yin2024improved,yin2024one,Salimans2024MultistepDO,Sauer2023AdversarialDD,lin2024sdxl,Sauer2024FastHI}. Although these methods differ technically, they share a core principle: the student is optimized with global objectives that match the teacher’s distributions or trajectories, rather than data-derived local velocities, so that it can perform few-step inference. A key limitation, however, is the \textit{reliance on a strong pretrained teacher}. This has recently motivated growing interest in self-contained, from-scratch training frameworks that natively yield few-step models.
A prominent line of work is consistency-based methods~\cite{song2023consistency,kim2023consistency,geng2025mean,lu2024simplifying,Sabour2025AlignYF,Boffi2024FlowMM,wang2025transition}, which essentially learn the underlying flow maps~\cite{Boffi2024FlowMM} -- or, equivalently, the average velocity~\cite{geng2025mean} between two points along the reverse trajectory -- so that, in principle, the model can follow a one-step shortcut instead of integrating many instantaneous velocities at test time. However, these objectives are typically unstable to optimize from scratch~\cite{imm,wang2025transition} or suffer from quality degradation~\cite{rcm}, and have so far scaled reliably only on simpler benchmarks such as ImageNet~\cite{deng2009imagenet}, while large text-to-image systems that do succeed in this regime still rely heavily on distillation~\cite{rcm,Sabour2025AlignYF}, undermining the original teacher-free motivation.

In this paper, we present the \textit{Self-Evaluating Model} (\textit{Self-E}, pronounced like selfie) -- a novel, self-contained, from-scratch training framework enabling any-step text-to-image inference.
The model learns simultaneously from data, which provides local velocity supervision, and through a novel \textit{self-evaluation} mechanism supervising the global distribution. The core idea is conceptually simple yet powerful: the model evaluates its own generated samples using its current local score estimate, effectively serving as a \emph{dynamic self-teacher}.
This self-evaluation becomes an increasingly accurate guidance signal -- allowing the model to improve itself.
By combining instantaneous local learning with self-driven global matching, \textit{Self-E} naturally bridges the gap between flow-based and distillation-based paradigms.
As a result, it can be trained entirely from scratch while supporting \emph{any-step} text-to-image inference, generating high-quality images even at very low step counts (see \cref{fig:teaser}).

To the best of our knowledge, \textit{Self-E} is the first native any-step text-to-image model, concurrent with TiM~\cite{wang2025transition}.
We conduct extensive experiments on large-scale text-to-image generation and show that \textit{Self-E} achieves both strong few-step quality and graceful scaling across inference budgets.
In the few-step ($<8$) setting, \textit{Self-E} surpasses the performance of diffusion and flow-based models including FLUX-1-dev~\cite{flux}, SDXL~\cite{podell2023sdxl}, SANA~\cite{xie2025sana}, the distillation-based LCM~\cite{luo2023latent}, and concurrent any-step model TiM~\cite{wang2025transition}.
Remarkably, although targeting this few-step generation, \textit{Self-E} is also competitive or even surpasses some state-of-the-art flow-based methods at the 50 step setting.
We also note that \textit{Self-E}’s performance improves monotonically as inference steps increase, enabling both ultra-fast few-step generation and high-quality long-trajectory sampling within a single unified model. 

\begin{figure*}
    \centering
    \includegraphics[width=\linewidth]{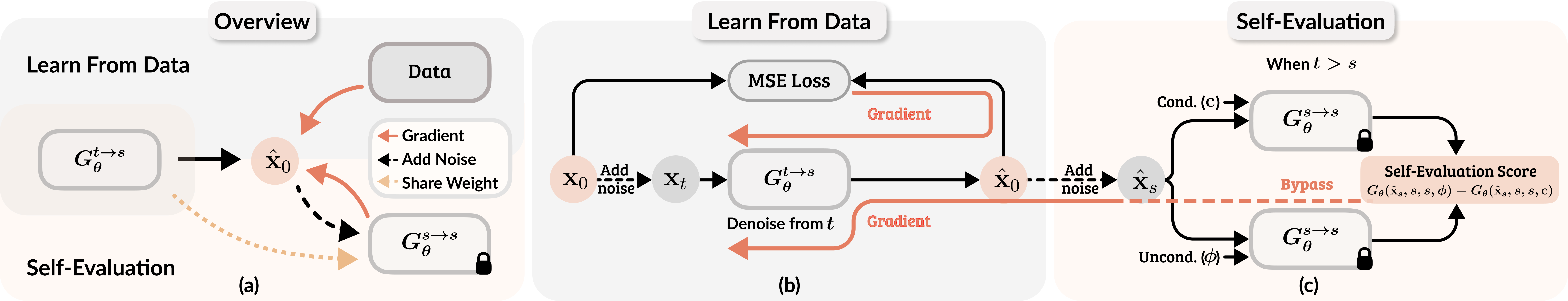}
    \caption{\textbf{Self-Evaluating Model.}
(a) \textit{Overview.} The model is trained with two complementary objectives: learning from data (b) and self-evaluation (c).
(b) \textit{Learning from data.} Given a real sample $\mathbf{x}_0$, we add noise to obtain $\mathbf{x}_t$ and train $G_\theta^{t \rightarrow s}$ with an $x_0$-prediction loss, providing local trajectory supervision.
(c) \textit{Self-evaluation with classifier score.} When $s<t$, we re-noise the generated $\hat{\mathbf{x}}_0$ to $\hat{\mathbf{x}}_s$ and run the same network in evaluation mode (stop-gradient) twice: once with condition $\mathbf{c}$ and once with the null prompt $\phi$. The difference between these outputs yields a self-evaluation score, which is treated as a feedback gradient on $\hat{\mathbf{x}}_0$ and back-propagated through the denoising path, enforcing global distribution matching in a teacher-free manner.
    }
    \label{fig:framework}
\end{figure*}

\section{Background: Flow Matching}
\label{sec:preliminary}

Generative models aim to learn a parameterized distribution $p_{\theta}(\mathbf{x}_0|\mathbf{c})$ that approximates the real data distribution $q(\mathbf{x}_0|\mathbf{c})$ where $\mathbf{c}$ is a condition such as text prompt. Flow matching and diffusion models achieve this by learning the instantaneous velocity field, or equivalently the score function, induced by a continuous-time forward diffusion process. Specifically, given real data samples $\mathbf{x}_0 \sim q(\mathbf{x}_0|\mathbf{c})$, these models define a trajectory indexed by $t \in [0,1]$:
\begin{equation}
\mathbf{x}_t = \alpha_t \mathbf{x}_0 + \sigma_t \boldsymbol{\epsilon}, \quad \boldsymbol{\epsilon} \sim \mathcal{N}(0, \mathbf{I}),
\end{equation}
where coefficients $(\alpha_t, \sigma_t)$ form a noise scheduler and defines the noisy distribution $q(\mathbf{x}_t|\mathbf{c})$. Flow matching specifically refers to a particular parameterization that explicitly matches the velocity field:
\begin{equation}
v_t(\mathbf{x}_t) := \frac{d\mathbf{x}_t}{dt} = \frac{d\alpha_t}{dt} \mathbf{x}_0 + \frac{d\sigma_t}{dt}\boldsymbol{\epsilon},
\end{equation}
where coefficients $(\alpha_t, \sigma_t) = (1 - t, t).$

Conditional Flow Matching (CFM) trains a neural network $V_{\theta}(\mathbf{x}_t, t, \mathbf{c})$ to predict the marginal velocity field (i.e., the expectation of instantaneous velocity)~\cite{flow_matching} by minimizing the mean squared error between predicted and conditional velocities:
\begin{equation}
\label{eq:velocity_matching_loss}
\mathcal{L}_{\text{CFM}}(\theta) = \mathbb{E}_{t, \mathbf{x}_0, \boldsymbol{\epsilon}}\left[\|V_{\theta}(\mathbf{x}_t, t, \mathbf{c}) - v_t(\mathbf{x}_t)\|^2\right].
\end{equation}

At inference time, we use the predicted velocity to follow the trajectory and generate samples. 
However, because the velocity is a \emph{local} quantity, a single-step estimate of the original sample $\mathbf{x}_0$ often falls short. 
Intuitively, a single step only captures the immediate direction and cannot account for the curvature of the trajectory, so it typically recovers just the \emph{average} of the possible original samples. 
Formally, a naive one-step estimate is
\begin{equation}
\label{eq:x0}
\hat{\mathbf{x}}_0 = \mathbf{x}_t - t\, V_{\theta^*}(\mathbf{x}_t, t, \mathbf{c}) \approx \mathbb{E}[\mathbf{x}_0|\mathbf{x}_t, \mathbf{c}],
\end{equation}
where $\theta^*$ minimizes \cref{eq:velocity_matching_loss}.

\section{Self-Evaluating Model}

We introduce the Self-Evaluating Model, a novel text-to-image pretraining approach enabling flexible, any-step inference. As illustrated in ~\cref{fig:framework}(a), the core idea is simple yet effective: the model simultaneously learns from data while performing self-evaluation. Conceptually, the loss function of our model is formulated as:

\begin{equation}
\label{eq:overall}
\mathcal{L}(\theta)= \mathcal{L}_{\text{data}}(\theta) + \lambda \mathcal{L}_{\text{self-evaluate}}(\theta).
\end{equation}

The learning-from-data component $\mathcal{L}_{\text{data}}(\theta)$ (\cref{sec:data}) provides local trajectory supervision, effectively estimating the conditional expectation $\mathbb{E}[\mathbf{x}_0|\mathbf{x}_t, \mathbf{c}]$. Meanwhile, the self-evaluation component $\mathcal{L}_{\text{self-evaluate}}$ (\cref{sec:self}) targets global distribution matching, encouraging the model-generated output $\hat{\mathbf{x}}_0$ to be a realistic sample drawn from the true distribution $q(\mathbf{x}_0)$. We demonstrate that surprisingly, this can be achieved through self-evaluation of the generated images by the model itself.

\paragraph{Model Parametrization.}
Formally, given a noisy input $\mathbf{x}_t$, we train a model $G_{\theta}(\mathbf{x}_t,t,s,\mathbf{c})$ to predict the clean data sample $\hat{\mathbf{x}}_0=G_\theta(\mathbf{x}_t,t,s,\mathbf{c})$, parameterized as:
\begin{equation}
\label{eq:our-x0}
    \hat{\mathbf{x}}_0 = G_\theta(\mathbf{x}_t,t,s,\mathbf{c})= \mathbf{x}_t - t\, \NetWork(\mathbf{x}_t, t,s,\mathbf{c}),
\end{equation}
where $\NetWork (\mathbf{x}_t, t,s,\mathbf{c})$ denotes a neural network analogous to $V_\theta$ from \cref{eq:velocity_matching_loss}, but noted distinctively as it takes in two time variables.
%
Our two time variables intuitively remind of self-consistency-based models~\cite{kim2023consistency,geng2025mean,frans2024one,Sabour2025AlignYF}, but here they serve a fundamentally different purpose.
Self-consistency methods essentially learn a specific underlying Flow Map~\cite{Boffi2024FlowMM} or an average velocity~\cite{geng2025mean} along the reverse trajectory, i.e., the integral of local velocities. In contrast, our goal is to directly predict samples whose marginal distribution \(p_\theta(\mathbf{x}_s|\mathbf{c})\) matches the real distribution \(q(\mathbf{x}_s|\mathbf{c})\), without constraining the reverse transition to follow any particular trajectory.

\subsection{Learning from Data}
\label{sec:data}

Our model is always trained on real data using the conditional flow matching loss in~\cref{eq:velocity_matching_loss}
which is equivalent to learning expectation of $\mathbf{x}_0$ prediction from $G_{\theta}(\mathbf{x}_t, t, s, \mathbf{c})$
through
\begin{equation}
\label{eq:data_loss}
\mathcal{L}_{\text{data}}(\theta) = \mathbb{E}_{s, t, \mathbf{x}_0, \boldsymbol{\epsilon}}\left[\|G_{\theta}(\mathbf{x}_t, t, s, \mathbf{c}) - \mathbf{x}_0\|^2\right].
\end{equation}

where $s \leq t$ are randomly sampled during training. In particular, when $s = t$, our model is optimized solely by this loss. The optimally trained $G_{\theta}(\mathbf{x}_t, t, t, \mathbf{c})$ serves as an estimate of the conditional expectation $\mathbb{E}[\mathbf{x}_0|\mathbf{x}_t,\mathbf{c}]$. However, the expectation itself may not be a meaningful sample in $q(\mathbf{x}_0|\mathbf{c})$.
Since this supervision is derived from the data distribution, we refer to this process as \emph{learning from data}.

\subsection{Learning by Self-Evaluation}

\label{sec:self}
When \(s < t\), we introduce another objective which targets at \textit{global distribution matching}. We interpret \(\hat{\mathbf{x}}_0 = G_{\theta}(\mathbf{x}_t,t,s,\mathbf{c})\) as a sample from an implicit distribution $p_{\theta}(\mathbf{x}_0|\mathbf{x}_t,t,s,\mathbf{c})$. Our goal is then to ensure that the marginal distribution:
\begin{equation}
\label{eq:margin}
p_\theta(\mathbf{x}_s|\mathbf{c})=\iint q(\mathbf{x}_s|\mathbf{x}_0)p_{\theta}(\mathbf{x}_0|\mathbf{x}_t,t,s,\mathbf{c})q(\mathbf{x}_t|\mathbf{c}) d\mathbf{x}_td\mathbf{x}_0
\end{equation}
closely matches the real distribution $q(\mathbf{x}_s|\mathbf{c})$. To accomplish this, we consider the reverse KL divergence between \(p_\theta(\mathbf{x}_s|\mathbf{c})\) and \(q(\mathbf{x}_s|\mathbf{c})\):
\begin{equation}
\label{eq:kl_reverse}
\begin{aligned}
&D_{\mathrm{KL}}\bigl(p_{\theta}(\mathbf{x}_s|\mathbf{c})\|q(\mathbf{x}_s|\mathbf{c})\bigr)
=\\
&\quad \mathbb{E}_{\mathbf{x}_s \sim p_{\theta}(\mathbf{x}_s|\mathbf{c})}\left[\log p_{\theta}(\mathbf{x}_s|\mathbf{c}) - \log q(\mathbf{x}_s|\mathbf{c})\right].
\end{aligned}
\end{equation}
The gradient of this KL divergence for per-sample optimization involves the difference between corresponding score functions:
\begin{equation}
    \label{eq:score_diff}
    \boldsymbol{\delta}(\hat{\mathbf{x}}_s) = \nabla_{\hat{\mathbf{x}}_s}\log p_{\theta}(\hat{\mathbf{x}}_s|\mathbf{c}) - \nabla_{\hat{\mathbf{x}}_s}\log q(\hat{\mathbf{x}}_s|\mathbf{c}),
\end{equation}
where we denote $\hat{\mathbf{x}}_s$ as a sample from $p_{\theta}({\mathbf{x}}_s|\mathbf{c})$.

\paragraph{Key Observation.}
Both score functions in \cref{eq:score_diff} are intractable in practice. Specifically, $\nabla_{\hat{\mathbf{x}}_s}\log q(\hat{\mathbf{x}}_s|\mathbf{c})$ represents the real-data score, \textit{which serves as the key driving force directing the sample towards regions of higher data density}. In contrast, $\nabla_{\hat{\mathbf{x}}_s}\log p_{\theta}(\hat{\mathbf{x}}_s|\mathbf{c})$ is termed the fake score, guiding the sample away from its current position and typically preventing mode collapse. To make optimization possible, prior methods use pre-trained diffusion to model real-data score, i.e., distill from a teacher model~\cite{yin2024one,yin2024improved,wang2023prolificdreamer}. \textit{We argue that, obtaining a perfect real score is unnecessary; instead, we leverage the currently trained model \(G_{\theta}(\mathbf{x}_s, s, s, \mathbf{c})\) to provide feedback for global distribution matching, which is a self-evaluation process.}

Formally, according to Tweedie's formula~\cite{efron2011tweedie, robbins1992empirical, chung2022improving}, the score function is related to the conditional expectation:
\begin{equation}
\label{eq:tweed}
    \nabla_{\mathbf{x}_s} \log q(\mathbf{x}_s|\mathbf{c}) = \frac{ \alpha_s \mathbb{E}[\mathbf{x}_0|\mathbf{x}_s,\mathbf{c}]-\mathbf{x}_s}{\sigma_s^2}.
\end{equation}

Note that our current model \(G_{\theta}(\mathbf{x}_s, s, s, \mathbf{c})\) progressively learns the expectation
\(\mathbb{E}[\mathbf{x}_0 | \mathbf{x}_s,\mathbf{c}]\) from the data, so we can use it to approximate the real score. Although this estimate is not fully accurate before convergence, it can still effectively guide training, since the “student” model itself is also far from converged in the early stages. Moreover, in practice the real score is typically evaluated under classifier-free guidance (CFG), and the reverse KL objective is inherently mode-seeking. Together, these properties provide stronger guidance for model optimization.

\paragraph{Self-Evaluation Score. } 
\label{sec:self-score}
We now concretely describe how we use the in-training model \(G_{\theta}(\mathbf{x}_s, s, s, \mathbf{c})\), which progressively learns the expectation
\(\mathbb{E}[\mathbf{x}_0 | \mathbf{x}_s,\mathbf{c}]\), to approximate the score terms in \cref{eq:score_diff} and \cref{eq:tweed}.
In common practice~\cite{yin2024one,yin2024improved,yu2023text}, the real score is evaluated within a conditionally sharpened distribution via classifier-free guidance (CFG)~\cite{ho2022classifier}, defined as:
\begin{equation}
\begin{aligned}
\nabla_{\hat{\mathbf{x}}_s} \log q_{w}(\hat{\mathbf{x}}_s|\mathbf{c}) 
&= \nabla_{\hat{\mathbf{x}}_s} \log q(\hat{\mathbf{x}}_s|\mathbf{c}) \\
&\quad + (\omega - 1)\nabla_{\hat{\mathbf{x}}_s} \log \frac{q(\hat{\mathbf{x}}_s|\mathbf{c})}{q(\hat{\mathbf{x}}_s|\phi)},
\end{aligned}
\end{equation}
where $\omega$ is a guidance scale, $\phi$ is a null prompt, denoting the unconditional distribution.
By subtracting the fake score and applying appropriate transformations, we rewrite \cref{eq:score_diff} into two distinct terms: which we call a classifier score term and an auxiliary term, i.e.:

\begin{equation}
\label{eq:score}
\begin{split}
\boldsymbol{\delta}(\hat{\mathbf{x}}_s)
=
(\omega-1)
\underbrace{
   \bigl(
    \nabla_{\hat{\mathbf{x}}_s} \log q(\hat{\mathbf{x}}_s|\phi)
    - \nabla_{\hat{\mathbf{x}}_s} \log q(\hat{\mathbf{x}}_s|\mathbf{c})
  \bigr)
}_{\text{Classifier score term}}
\\[4pt]
\quad+\;
\underbrace{
  \bigl(
    \nabla_{\hat{\mathbf{x}}_s}\log p_{\theta}(\hat{\mathbf{x}}_s | \mathbf{c})
    - \nabla_{\hat{\mathbf{x}}_s} \log q(\hat{\mathbf{x}}_s|\mathbf{c})
  \bigr)
}_{\text{Auxiliary term (optional)}}.
\end{split}
\end{equation}

Empirically, we observe that using only the classifier score term is sufficiently effective and even improves convergence (see \cref{tab:ablation}). This observation is consistent with prior work~\cite{yu2023text}, which also found classifier scores effective when performing score distillation for 3D generation~\cite{poole2022dreamfusion}. Consequently, we omit the auxiliary term and thereby avoid co-training an additional model for the fake score during the early stages of training. Although this setting no longer corresponds to exact distribution matching, it still provides a meaningful learning signal: intuitively, the classifier score encourages the model to generate samples that align with an implicit classifier $q(\mathbf{c}|\mathbf{x})$~\cite{yu2023text,ho2022classifier}. 

The fake score primarily helps prevent mode collapse; in our case, we note that the learning-from-data component can already fulfill this role. When the model is close to convergence, i.e., in later training stages, we can optionally re-introduce the auxiliary term to perform more accurate distribution matching, which helps reduce artifacts (see \cref{fig:csd_vs_vsd}). Even then, we do not require an additional copy of the model; instead, we simply utilize a specialized prompt to estimate the generated score. We provide more details about this case in the supplementary material.

We now formally describe our practical implementation of the self-evaluation score using the in-training network.  
The detailed procedure of self-evaluation with only the classifier score is illustrated in \cref{fig:framework}(c).
We employ two stop-gradient forward passes with self-generated samples as input.
In particular, we add noise to the generated sample $\hat{\mathbf{x}}_0$: $\hat{\mathbf{x}}_s = \alpha_s \hat{\mathbf{x}}_0 + \sigma_s \boldsymbol{\epsilon}$, and define a pseudo-target as:
\begin{equation}
    \mathbf{x}_{\text{self}} := \mathrm{sg}[\hat{\mathbf{x}}_0 - [G_{\theta}(\hat{\mathbf{x}}_s, s, s, \phi)-G_{\theta}(\hat{\mathbf{x}}_s, s, s, \mathbf{c})]],
\end{equation}
where $\mathrm{sg}$ denotes the stop-gradient operation. Minimizing the mean squared error (MSE) between this pseudo-target and our model's prediction induces a gradient with respect to $\hat{\mathbf{x}}_s$ that precisely matches the desired direction, i.e., the classifier score in \cref{eq:score}. We provide the proof in the appendix. Thus, our self-evaluation loss is expressed as: 
\begin{equation}
    \label{eq:se_loss}
    \mathcal{L}_{\text{self-evaluate}}(\theta) = \mathbb{E}_{t, s, \mathbf{x}_0, \boldsymbol{\epsilon}}\left[\|G_{\theta}(\mathbf{x}_t, t, s, \mathbf{c}) - \mathbf{x}_{\text{self}}\|^2\right].
\end{equation}

\subsection{Final Objective}

Our final per-sample objective is a hybrid loss function that combines data-driven supervision with global distribution matching via self-evaluation. Formally, it is defined as:
\begin{equation}
\label{eq:unify_loss}
    \mathcal{L}_{s,t}(\theta) = 
    \|\hat{\mathbf{x}}_0 - \mathbf{x}_0\|_2^2
    + \lambda_{s,t}\,\|\hat{\mathbf{x}}_0 - \mathbf{x}_{\text{self}}\|_2^2,
\end{equation}
where the weight $\lambda_{s,t}$ controls the relative contribution of the self-evaluation term and is given by:
\begin{equation}
    \lambda_{s,t} = \frac{\sigma_t}{\alpha_t} - \frac{\sigma_s}{\alpha_s}.
\end{equation}
Note that $\lambda_{s,t} = 0$ when $t = s$, in which case the objective reduces to a purely data-driven reconstruction loss.

In practice, large values of $\lambda_{s,t}$ can overpower the data-driven loss, leading to undesired color bias.
To mitigate this effect, we introduce an \emph{energy-preserving normalization} of the effective training target, inspired by~\citet{zhang2024ep}, which addresses a similar issue with high CFG values. 
From the gradient of \cref{eq:unify_loss}, the implicit regression target can be expressed as:
\begin{equation}
    \mathbf{x}_{\text{tar}} = \frac{\mathbf{x}_0 + \lambda_{s,t} \mathbf{x}_{\text{self}}}{1 + \lambda_{s,t}}.
\end{equation}
We normalize this target to preserve the energy of the clean sample $\mathbf{x}_0$:
\begin{equation}
\label{eq:final_renorm}
    \mathbf{x}_{\mathrm{renorm}} = \frac{\mathbf{x}_0 + \lambda_{s,t} \mathbf{x}_{\text{self}}}{\|\mathbf{x}_0 + \lambda_{s,t} \mathbf{x}_{\text{self}}\|_2} \|\mathbf{x}_0\|_2.
\end{equation}
Empirically, this normalization yields slightly improved visual quality and stability (see \cref{tab:ablation}). 
Replacing $\mathbf{x}_{\text{tar}}$ with $\mathbf{x}_{\mathrm{renorm}}$, our practical per-pair objective becomes:
\begin{equation}
\label{eq:renorm_loss}
    \mathcal{L}_{s,t}(\theta) = \|\hat{\mathbf{x}}_0 - \mathbf{x}_{\mathrm{renorm}}\|_2^2.
\end{equation}

Finally, the overall training loss is obtained by averaging over all possible timestep pairs:
\begin{equation}
    \mathcal{L}(\theta) = \mathbb{E}_{s,t}\left[w_{s,t}\,\mathcal{L}_{s,t}(\theta)\right],
\end{equation}
where $w_{s,t}$ denotes the sampling weight for each pair $(s,t)$.

\subsection{Inference}
\label{sec:inference}

Our model supports inference with an arbitrary number of steps by iteratively removing noise, similar to diffusion and flow matching models. Given a predefined inference step budget $N$ and a corresponding time scheduler $\{t_{k}\}$, where $1 = t_1 > t_2 > \dots > t_N = 0$, we sequentially predict a denoising direction at each timestep $t_k$ and take a step towards the next timestep $t_{k+1}$. Formally, each inference step is defined as:
\begin{align}
\mathbf{x}_{t_{k+1}} = \mathbf{x}_{t_k} - (t_k - t_{k+1}) \NetWork (\mathbf{x}_{t_k}, t_k, s_k, \mathbf{c}).
\end{align} 

By default, we set the target timestep $s_k$ to be the next timestep $t_{k+1}$.
Nevertheless, we find that setting the timestep $s_k$ to other values in the interval $[t_{k+1}, t_k]$ might lead to improved results in some cases. We demonstrate this phenomenon and provide some suggestions to setting this hyperparameter in the supplementary material. 
We employ energy-preserving classifier-free guidance~\cite{zhang2024ep} with $\omega=5$.

\section{Experiment}
\begin{figure*}[ht]
  \centering
  \includegraphics[width=\textwidth]{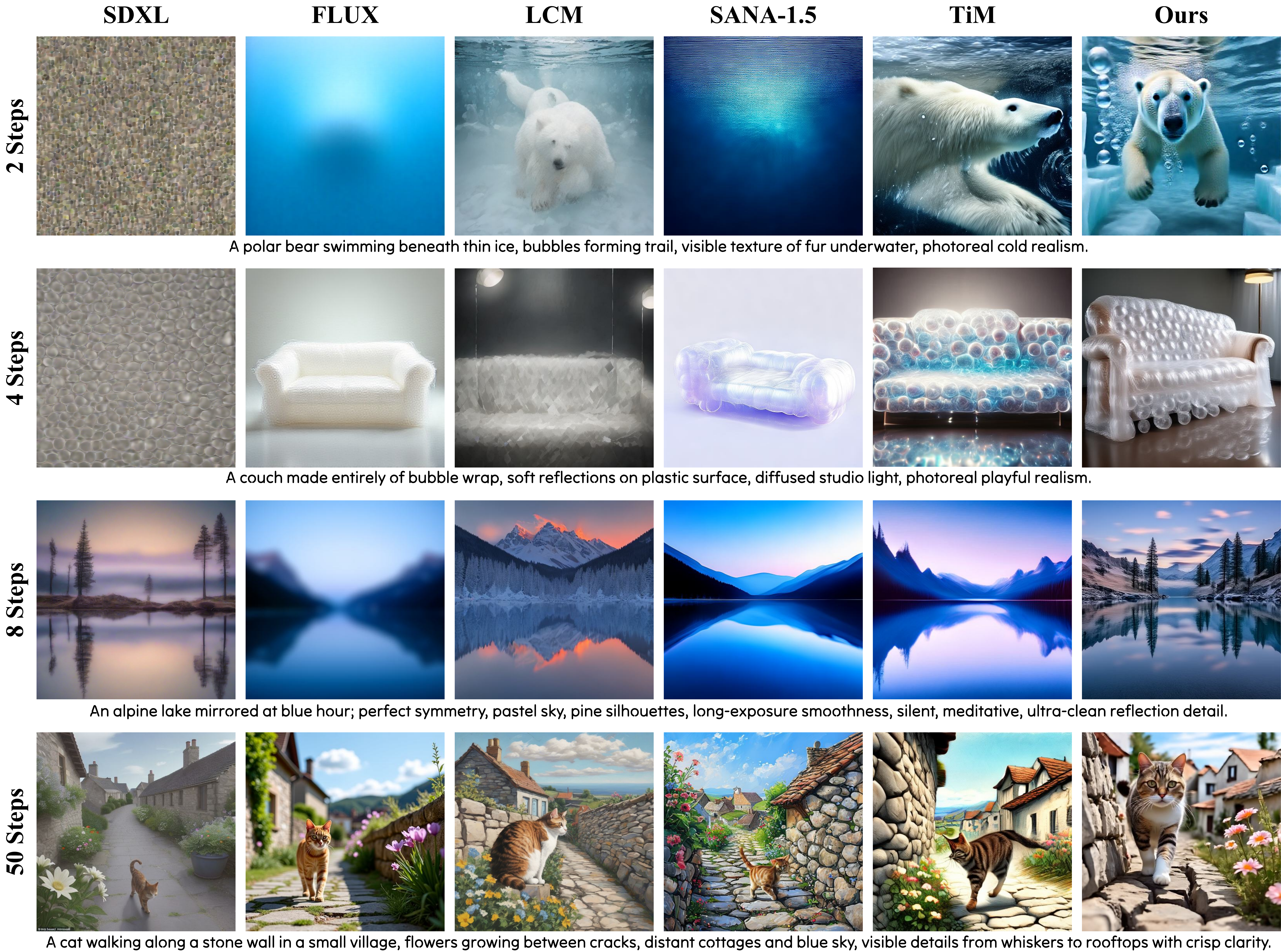}
  \caption{\textbf{Qualitative Any-Step Comparison.} Generated images from all methods at various inference steps. Our approach consistently produces detailed, semantically accurate, and visually appealing images aligned with textual prompts at all step counts. In extremely few-step scenarios (e.g., 2-step), FLUX, SANA, and SDXL fail to generate recognizable results, while LCM and TiM exhibit semantic and structural degradation. 
  When using more inference steps, all methods improve, but our method retains superior quality, realism, and text alignment.
  At 50 steps, normal Flow Matching realm, our method is competitive with FLUX, despite FLUX being a much larger model.}
  \label{fig:compare}
\end{figure*}
\begin{table*}[t]
\centering
\caption{\textbf{Quantitative Comparison on GenEval~\cite{geneval}.} Our method is \emph{consistently SOTA} across all step counts and improves monotonically with more steps on \textit{GenEval Overall} (2$\rightarrow$4$\rightarrow$8$\rightarrow$50: 0.753$\rightarrow$0.781$\rightarrow$0.785$\rightarrow$0.815). Notably, we achieve \textbf{large margins in the few-step regime} (e.g., \textbf{+0.12} at 2-step over the best prior methods), while remaining the top performer at 8 and 50 Steps.}
\footnotesize
\renewcommand{\arraystretch}{1.15}
\setlength{\tabcolsep}{4pt}
\resizebox{0.99\textwidth}{!}{%
\begin{tabular}{l c | c c c c c c c}

\toprule
{\textbf{ Steps}} & \textbf{Method} & \textbf{Overall} $\uparrow$ & \textbf{Single Object} $\uparrow$ & \textbf{Two Object} $\uparrow$ & \textbf{Attribute Binding} $\uparrow$ & \textbf{Colors} $\uparrow$ & \textbf{Counting} $\uparrow$ & \textbf{Position} $\uparrow$\\
\midrule

\multirow{6}{*}{2 Steps}
& SDXL~\cite{podell2023sdxl} & 0.0021 & 0.0130 & 0.0000 & 0.0000 & 0.0000 & 0.0000 & 0.0000 \\
& FLUX.1-Dev~\cite{flux} & 0.0998 & 0.2969 & 0.0227 & 0.0025 & 0.1835 & 0.0656 & 0.0275 \\
& LCM~\cite{luo2023latent} & 0.2624 & 0.7937 & 0.0985 & 0.0050 & 0.4761 & 0.1812 & 0.0200 \\
& SANA-1.5~\cite{xie2025sana} & 0.1662 & 0.5531 & 0.0707 & 0.0075 & 0.2234 & 0.1125 & 0.0030 \\
& TiM~\cite{wang2025transition} & 0.6338 & 0.9469 & 0.7071 & 0.4375 & 0.8723 & 0.4188 & 0.4200 \\
& SDXL-Turbo~\cite{sdxl-turbo} & 0.4622 & 0.9781 & 0.3308 & 0.1500 & 0.7527 & 0.4594 & 0.1025 \\
& SD3.5-Turbo~\cite{sd3.5-turbo} & 0.3635 & 0.7125 & 0.2879 & 0.1650 & 0.5691 & 0.2812 & 0.1650 \\
\rowcolor{highlightcolor}
& Ours & \textbf{0.7531} & 0.9812 & 0.8838 & 0.5900 & 0.8218 & 0.6094 & 0.6325 \\
\midrule
\multirow{6}{*}{4 Steps}
& SDXL~\cite{podell2023sdxl} & 0.1576 & 0.5281 & 0.0758 & 0.0125 & 0.2606 & 0.0437 & 0.0250 \\
& FLUX.1-Dev~\cite{flux} & 0.3198 & 0.6469 & 0.2955 & 0.0550 & 0.4202 & 0.2437 & 0.2575 \\
& LCM~\cite{luo2023latent} & 0.3277 & 0.9344 & 0.1667 & 0.0150 & 0.5372 & 0.2656 & 0.0475 \\
& SANA-1.5~\cite{xie2025sana} & 0.5725 & 0.9219 & 0.6313 & 0.2525 & 0.6968 & 0.5125 & 0.4200 \\
& TiM~\cite{wang2025transition} & 0.6867 & 0.9531 & 0.7601 & 0.5225 & 0.9016 & 0.5031 & 0.4800 \\
& SDXL-Turbo~\cite{sdxl-turbo} & 0.4766 & 0.9781 & 0.4040 & 0.1400 & 0.7713 & 0.4562 & 0.1100 \\
& SD3.5-Turbo~\cite{sd3.5-turbo} & 0.7194 & 0.9344 & 0.8510 & 0.5650 & 0.7952 & 0.5656 & 0.6050 \\
\rowcolor{highlightcolor}
& Ours & \textbf{0.7806} & 0.9688 & 0.9141 & 0.6250 & 0.8936 & 0.6219 & 0.6600 \\
\midrule
\multirow{6}{*}{8 Steps}
& SDXL~\cite{podell2023sdxl} & 0.3759 & 0.8812 & 0.2702 & 0.0675 & 0.6569 & 0.2594 & 0.1200 \\
& FLUX.1-Dev~\cite{flux} & 0.5893 & 0.8844 & 0.7298 & 0.2175 & 0.7314 & 0.4625 & 0.5100 \\
& LCM~\cite{luo2023latent} & 0.3398 & 0.9281 & 0.1818 & 0.0300 & 0.5319 & 0.3094 & 0.0575 \\
& SANA-1.5~\cite{xie2025sana} & 0.7788 & 0.9812 & 0.8864 & 0.5800 & 0.9202 & 0.6750 & 0.6300 \\
& TiM~\cite{wang2025transition} & 0.7143 & 0.9656 & 0.8232 & 0.5750 & 0.8936 & 0.5156 & 0.5125 \\
& SDXL-Turbo~\cite{sdxl-turbo} & 0.4652 & 0.9688 & 0.3763 & 0.1300 & 0.7500 & 0.4562 & 0.1100 \\
& SD3.5-Turbo~\cite{sd3.5-turbo} & 0.7071 & 0.9437 & 0.8232 & 0.5450 & 0.8271 & 0.5312 & 0.5725 \\
\rowcolor{highlightcolor}
& Ours & \textbf{0.7849} & 0.9688 & 0.9141 & 0.6225 & 0.8830 & 0.6688 & 0.6525 \\
\midrule
\multirow{6}{*}{50 Steps}
& SDXL~\cite{podell2023sdxl} & 0.4601 & 0.9688 & 0.4217 & 0.1300 & 0.8138 & 0.3312 & 0.0950 \\
& FLUX.1-Dev~\cite{flux} & 0.7966 & 0.9781 & 0.9318 & 0.5600 & 0.9096 & 0.7500 & 0.6500 \\
& LCM~\cite{luo2023latent} & 0.3303 & 0.8938 & 0.2247 & 0.0075 & 0.5319 & 0.2812 & 0.0425 \\
& SANA-1.5~\cite{xie2025sana} & 0.8062 & 0.9844 & 0.9192 & 0.7175 & 0.9229 & 0.7031 & 0.5900 \\
& TiM~\cite{wang2025transition} & 0.7797 & 0.9656 & 0.8864 & 0.7300 & 0.9069 & 0.6344 & 0.5550 \\
& SDXL-Turbo~\cite{sdxl-turbo} & 0.3983 & 0.9156 & 0.2980 & 0.0700 & 0.6702 & 0.3563 & 0.0800 \\
& SD3.5-Turbo~\cite{sd3.5-turbo} & 0.6114 & 0.8656 & 0.7449 & 0.4050 & 0.6995 & 0.4281 & 0.5250 \\
\rowcolor{highlightcolor}
& Ours & \textbf{0.8151} & 0.9875 & 0.9394 & 0.6700 & 0.8910 & 0.7000 & 0.7025 \\
\bottomrule
\end{tabular}}
\label{tab:geneval}
\end{table*}

\label{sec:exp}

We conduct two complementary sets of experiments to validate our approach.
First, we train a 2B-parameter model with $512\times512$-resolution images and compare against state-of-the-art text-to-image models spanning the landscape of training paradigms (\cref{subsec:sota}).
Second, we perform controlled ablation studies with 0.5B-parameter models under identical training conditions to isolate the contributions of key design choices (\cref{subsec:ablation}).
In both, we adopt a latent transformer architecture similar to FLUX~\cite{flux, esser2024scaling}, with minor modifications to accommodate the additional timestep input $s$, which mirrors the typical handling of timestep input $t$.
Additional details about architecture, data,  and hyperparameters are provided in the supplementary material.

\subsection{Comparison with Prior Work}
\label{subsec:sota}

We compare our method with several state-of-the-art text-to-image approaches spanning the landscape of training paradigms. 
Most closely related to our setting are \emph{from-scratch any-step methods}. Since all published works in this category have been demonstrated only on small-scale datasets, such as CIFAR10~\cite{cifar10} and ImageNet~\cite{imm,geng2025mean}, we compare our model with the concurrent Transition Models (TiM)~\cite{wang2025transition}, which are the first to scale this family of approaches to text-to-image generation. 
In addition, we include standard flow-matching and diffusion baselines -- FLUX.1-dev~\cite{flux}, SDXL~\cite{podell2023sdxl}, and SANA-1.5~\cite{xie2025sana}. 
Finally, we compare with Latent Consistency Models (LCM)~\cite{luo2023latent}, SDXL-Turbo~\cite{sdxl-turbo}, and SD3.5-Turbo~\cite{sd3.5-turbo}, which employ different distillation methods from the pretrained Stable Diffusion model~\cite{rombach2022high} for few-step sampling. 
Note that these models, as well as other distillation-based approaches~\cite{yin2024one,yin2024improved}, are not trained from scratch and require a pretrained teacher.

Following the evaluation protocol of \citet{deng2025bagel}, we report quantitative results on the GenEval benchmark~\cite{geneval}. 
As shown in \cref{tab:geneval}, our method consistently outperforms other methods across all inference step counts, achieving notably higher scores overall. 
In the few-step regime, our model outperforms the second-best method by a large margin.
In \cref{fig:compare}, we visualize generated images from representative methods across 2, 4, 8, and 50 inference steps. 
Our approach consistently produces high-quality, detailed, and text-coherent images at all step counts. 
In the extreme few-step setting (2 steps), FLUX, SANA, and SDXL fail to generate meaningful images, while LCM and TiM produce recognizable objects but suffer from significant degradation in structure and semantic coherence. 
In contrast, our method yields clear, semantically aligned, and visually detailed results even under this challenging configuration. 
As the number of inference steps increases to 4 and 8 steps, all methods progressively improve, yet our approach maintains a clear advantage in both detail and text alignment. 
At 50 steps, our model attains image quality comparable or better than SANA, and SDXL, while LCM and TiM exhibit saturation artifacts.

\subsection{Ablation Studies}
\label{subsec:ablation}

To isolate the advantages of our approach over alternatives and to assess the effects of key design choices, we conduct controlled ablation experiments.
We use 0.5B-parameter models trained on identical datasets under consistent conditions, enabling direct comparison without confounding factors.
All ablations use $256 \times 256$ resolution and batch size 1024. Other settings follow the setup in \cref{subsec:sota} and are detailed in the supplementary material.

\paragraph{Comparison with Pretraining Alternatives.}

\begin{figure}[t]
  \centering
  \includegraphics[width=0.97\linewidth]{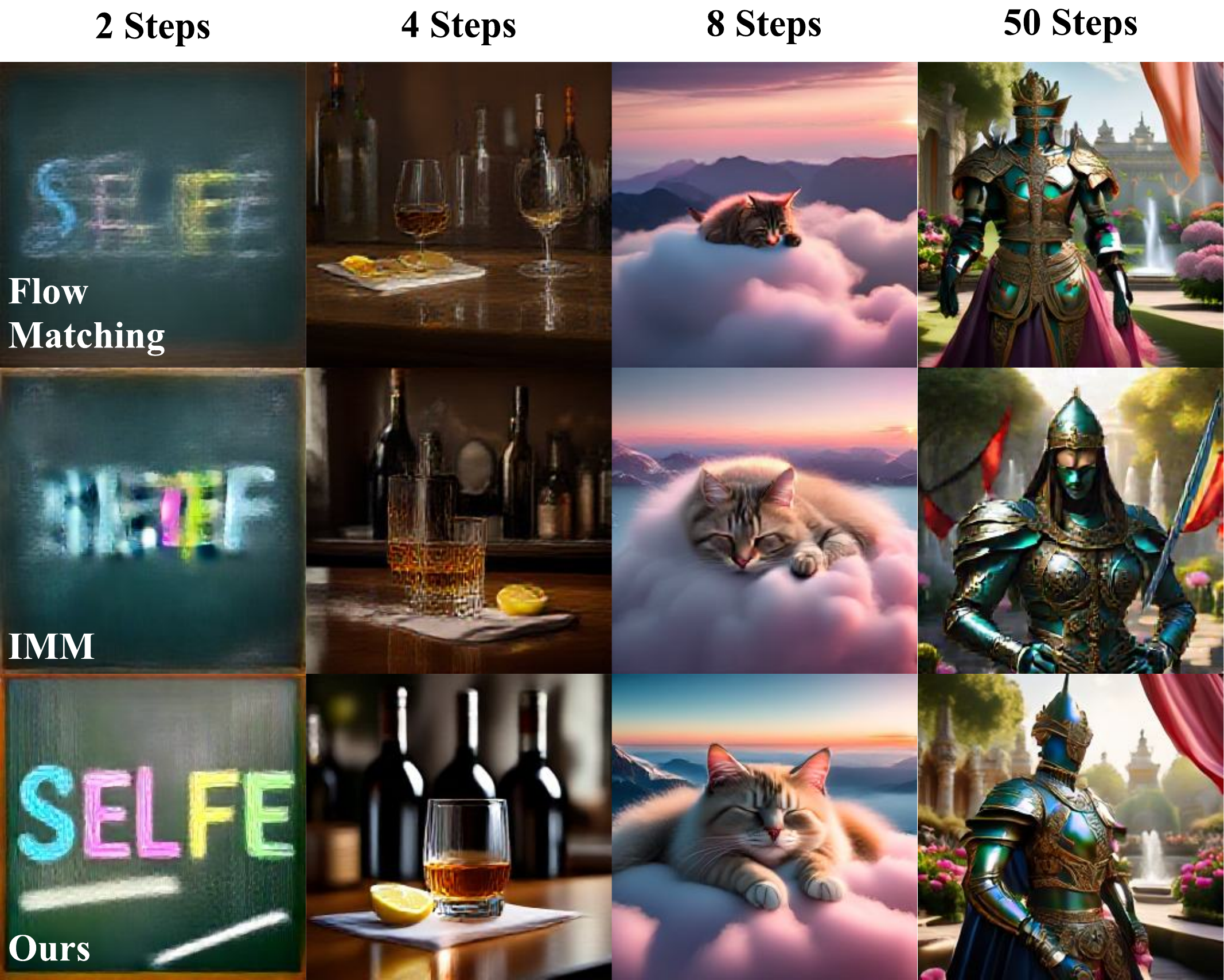}
  \caption{\textbf{Controlled Ablation Study.} We compare our method to alternative pretraining methods - Flow Matching and IMM. Full prompts appear in supplementary.
  Our method produces favorable results across all step budgets.
  %
  }
  \label{fig:ablation}
\end{figure}

\begin{figure}[t]
  \includegraphics[width=0.95\linewidth]{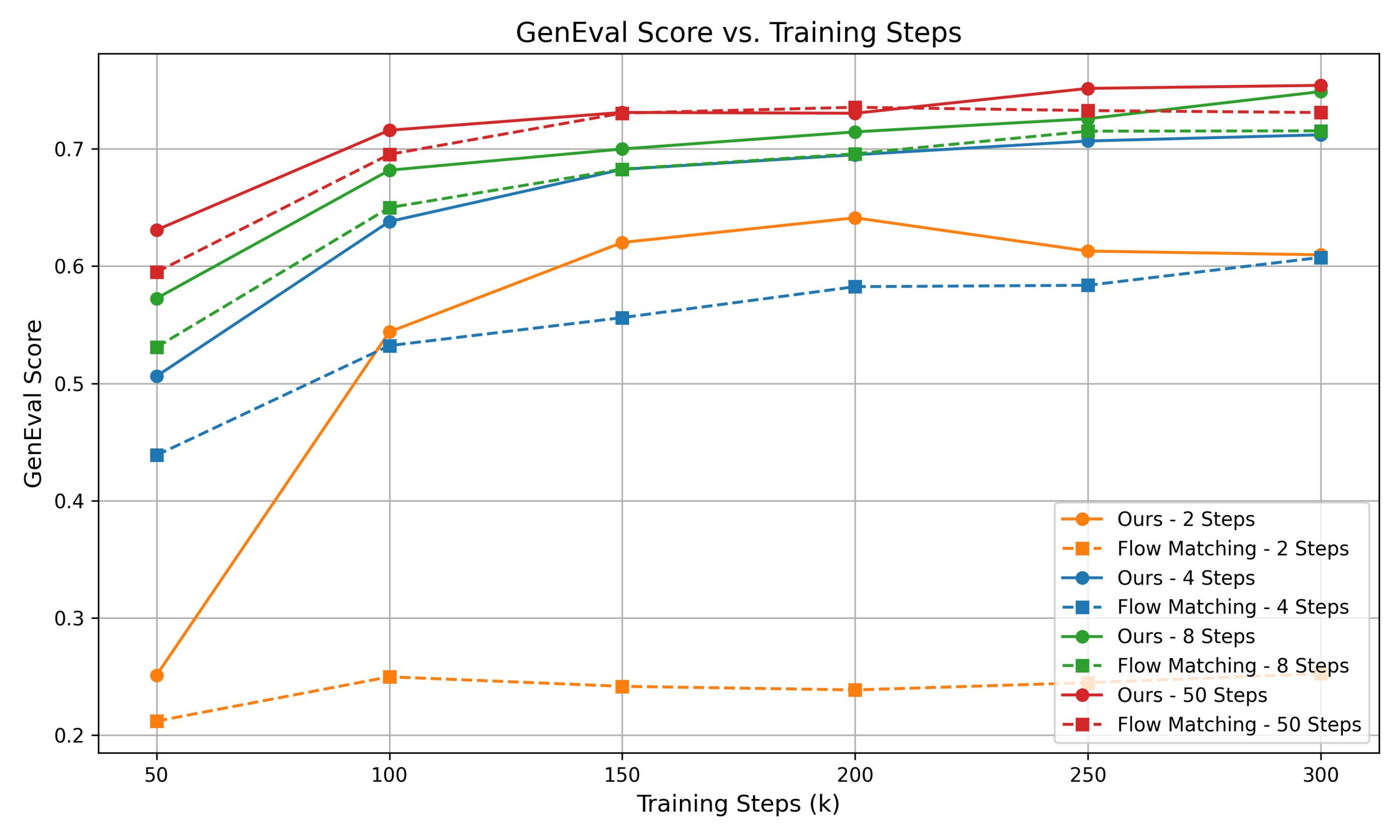}
  \caption{\textbf{Training Progress Comparison.} GenEval scores across different inference steps (2, 4, 8, and 50) for our method and Flow Matching over training iterations (from 50k to 300k). Our approach consistently outperforms Flow Matching at all inference steps, indicating its superior effectiveness and robustness.}
  \label{fig:geneval}
\end{figure}

Here we compare our approach with two paradigms: standard Flow Matching and native few-step methods.
For native few-step methods, we choose to experiment with a second method of this family -- the recently proposed, Inductive Moment Matching (IMM)~\cite{imm}, as our baseline.
IMM can be viewed as an extension of trajectory-based models to the distribution level via moment matching.
We report GenEval results in~\cref{tab:ablation} and provide corresponding qualitative comparisons in Figure~\ref{fig:ablation}. Our approach consistently outperforms both Flow Matching and IMM across all step counts. In Figure~\ref{fig:geneval}, we further plot GenEval scores for our method and Flow Matching throughout training, clearly demonstrating that our approach not only converges to superior performance but also maintains this advantage throughout the entire training.

\begin{figure}[t]
  \centering
  \includegraphics[width=0.97\linewidth]{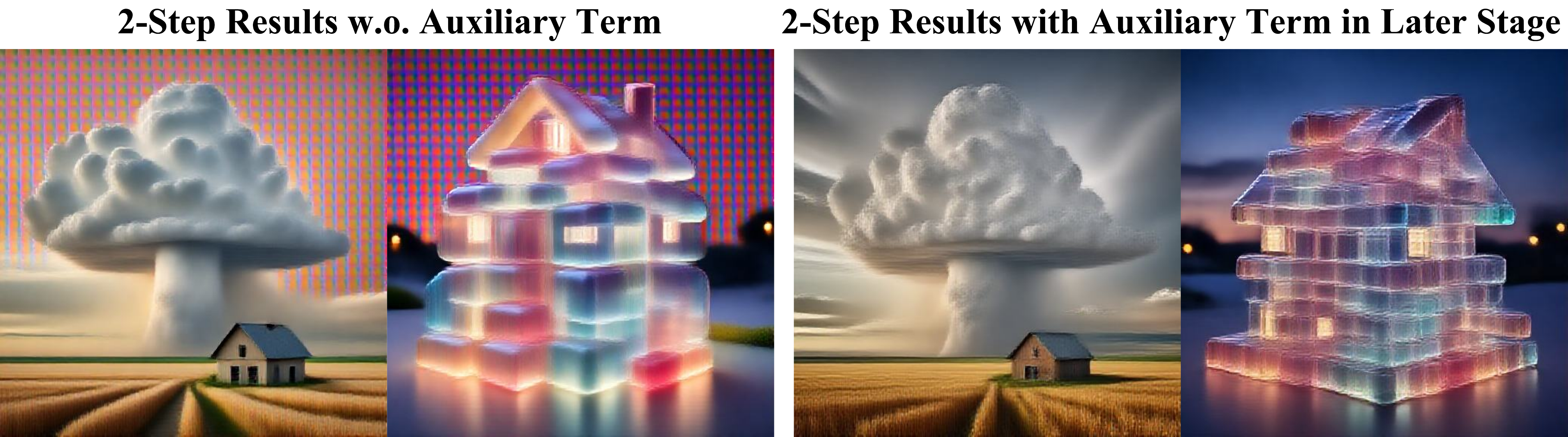}
  \caption{
  \textbf{(Left)} Models trained only with the classifier score component from~\cref{eq:score} have clear checkerboard artifacts in extreme few-step regime, 2 steps in this example. \textbf{(Right)} Incorporating the auxiliary term from~\cref{eq:score} in later stages of training helps mitigating these artifacts.
  Results are from our 2B model.
  }
  \label{fig:csd_vs_vsd}
\end{figure}

\paragraph{Design Choices.} We further investigate two design choices and analyze their individual effects by training each variant for 100k iterations.
In \cref{tab:ablation} we present results comparing models trained without energy-preserving target normalization (i.e. using \cref{eq:unify_loss} instead of \cref{eq:renorm_loss}) and models trained with the auxiliary term from \cref{eq:score} included throughout all training iterations.
We observe that target normalization generally improves performance, except in the extreme two-step inference setting, so we adopt this strategy for all our experiments. In contrast, introducing the auxiliary term from scratch significantly degrades performance. Therefore, we rely primarily on the classifier score in the early stages of training, which both reduces computational cost and stabilizes optimization. However, we find that incorporating the auxiliary term in later training stages is beneficial, notably mitigating oversaturated stripe artifacts in two-step generations, as shown in \cref{fig:csd_vs_vsd}. Consequently, we adopt this hybrid schedule -- classifier-score-only early, with including auxiliary term for refinement later -- as our final training strategy in the main experiments.

\begin{table}[t]
  \centering
  \caption{
  \textbf{Controlled Ablation Study.} We report overall scores on GenEval~\cite{geneval}. 
  The upper block compares our method with two alternative design choices of omitting the target normalization or incorporating the auxiliary term throughout all training steps. Reported after 100K iterations. 
  The bottom block compares our method with alternative pretraining methods - Flow Matching and IMM. Reported after 300K iterations.
  }
  \label{tab:ablation}
  \small
  \setlength{\tabcolsep}{6pt}
  \renewcommand{\arraystretch}{1.12}
  \resizebox{\columnwidth}{!}{%
  \begin{tabular}{lcccc}
    \toprule
    Method & 2 Steps $\uparrow$ & 4 Steps $\uparrow$ & 8 Steps $\uparrow$ & 50 Steps $\uparrow$ \\
    \midrule
    \multicolumn{5}{l}{\textit{100k Iterations}} \\
    w/o target norm.          & 0.5555 & 0.6156 & 0.6521 & 0.7018 \\
    w/ aux. term             & 0.3307 & 0.4304 & 0.5153 & 0.6166 \\
    \rowcolor{highlightcolor}
    \textbf{Ours}                   & 0.5439 & 0.6381  & 0.6819 & 0.7160 \\
    \midrule
    \multicolumn{5}{l}{\textit{300k Iterations}} \\
    Flow Matching~\cite{flow_matching}                  & 0.2523 & 0.6075  & 0.7155 & 0.7311 \\
    IMM~\cite{imm}                   & 0.2617 & 0.5994  & 0.7112 & 0.7472 \\
    \rowcolor{highlightcolor}
    \textbf{Ours}                   & 0.6097 & 0.7121  & 0.7490 & 0.7543 \\
    \bottomrule
  \end{tabular}%
  }
\end{table}

\section{Related Work}
\label{sec:work}

\noindent \textbf{Diffusion and Flow Matching.} 
Diffusion models \citep{sohl2015deep, song2019generative, ho2020denoising, song2020score} and flow-matching models \citep{flow_matching, albergo2022building, lipman2022flow, liu2022flow} have become two of the most popular frameworks for generative modeling in recent years. These models are trained to learn either a score function or a velocity field that reverses a noising process, transporting samples from the clean data distribution back to a simple prior distribution such as a Gaussian. Both diffusion and flow-matching approaches have been successfully scaled to a wide range of generative tasks, including text-to-image synthesis~\citep{chen2023pixart, podell2023sdxl, rombach2022high, zhou2024transfusion, esser2024scaling, flux}, text-to-video generation~\citep{blattmann2023stable, kong2024hunyuanvideo, gao2025seedance, yang2024cogvideox, opensora2, wan2025wan}, and large language modeling~\citep{nie2025large}. Despite their impressive performance, diffusion and flow-matching models are fundamentally designed to predict local properties of the data distribution. As a result, they typically require many iterative denoising steps to produce high-quality samples, which can pose significant computational challenges during inference.

\noindent \textbf{Accelerating Diffusion/Flow Matching.} 
There has been a rich body of literature focused on reducing the number of denoising steps required by diffusion and flow matching. Training-free approaches typically employ high-order solvers to better approximate the underlying differential equations \citep{lu2022dpm, karras2022elucidating, dockhorn2022genie, zhang2022fast, sabour2024align, zheng2023dpm, lu2025dpm}, but these methods still struggle to achieve high-quality samples within ten denoising steps. Another major line of work aims to accelerate diffusion models through distillation. Early distillation techniques train a student model to match the long-step transitions along the trajectory produced by a multi-step teacher \citep{salimans2022progressive, flow_matching}.
Consistency Models (CMs) and their variants \citep{song2023consistency, geng2024consistency, lu2024simplifying, song2023improved} instead learn a direct flow map that transports a noisy input directly to its corresponding clean sample by following the PF-ODE trajectory. Flow-map models further generalize this paradigm by learning mappings between arbitrary pairs of points (s, t) along the PF-ODE trajectory \citep{kim2023consistency, Zheng2024TrajectoryCD, frans2024one, Sabour2025AlignYF, Hu2025CMTMF, Boffi2024FlowMM, Heek2024MultistepCM, Wang2024PhasedCM}. More recent work, such as TiM \citep{wang2025transition} and MeanFlow \citep{geng2025mean}, attempts to learn such flow-map models through large-scale pre-training; however, we observe that these techniques remain difficult to scale effectively to text-to-image generation.

Another approach to obtain few-step models is distribution-matching distillation \citep{yin2024improved, yin2024one, Salimans2024MultistepDO, Sauer2024FastHI, Sauer2023AdversarialDD, Zhou2024AdversarialSI, Zhou2024ScoreID}, where different divergence metrics are employed as training losses and applied to samples at different noise levels to move student generated sample towards teacher's learned distribution. Our work is inspired by the distribution-matching viewpoint, but differs in that we apply this idea during the pre-training stage of text-to-image models.

\section{Conclusion}
\label{sec:conclusion}

In this paper, we introduce the \textit{Self-Evaluating Model (Self-E)}, a novel pretraining framework for text-to-image generation capable of flexible, any-step inference entirely from scratch. Departing from prior approaches dependent on pretrained teacher models, \textit{Self-E} leverages dynamically learned local scores to self-assess generated samples, establishing an internal feedback loop that seamlessly integrates local trajectory learning with global distribution matching. Comprehensive evaluations on the GenEval benchmark demonstrate \textit{Self-E}’s state-of-the-art performance across diverse inference budgets, particularly excelling in few-step generation scenarios. Furthermore, \textit{Self-E}’s performance monotonically improves with increased inference steps, indicating its capability to scale from rapid generation to high-quality long-trajectory sampling. We hope \textit{Self-E} offers a fresh perspective on designing teacher-free pretraining methods for any-step image generation and inspires future work on transferring such self-evaluating models to downstream tasks.


{
    \small
    \bibliographystyle{ieeenat_fullname}
    \bibliography{main}
}


\clearpage
\begin{center}
    {\LARGE \textbf{Appendix}}
\end{center}

\vspace{8pt}

This appendix provides additional details and results complementing the main paper. In \cref{sec:s1-self-eval}, we provide proofs showing that our proposed self-evaluation loss correctly induces the desired optimization gradients. We distinguish two scenarios: deriving the classifier-score gradient and deriving the full reverse KL divergence gradient. For the latter, we include additional implementation details on training. \cref{sec:s2-implementation} contains extended information on our training and inference implementation. In \cref{sec:s3-experiments}, we present additional experimental results and further discussions regarding the choice of the second timestep input. Prompts corresponding to image examples shown in the main paper are provided in \cref{sec:s4-prompt}. Finally, in \cref{sec:s5-limitation}, we discuss limitations of our method and propose directions for future work.

\vspace{12pt}

\setcounter{section}{0}
\setcounter{subsection}{0}
\setcounter{subsubsection}{0}
\renewcommand{\thesection}{S.\arabic{section}}
\renewcommand{\thesubsection}{S.\arabic{section}.\arabic{subsection}}
\renewcommand{\thesubsubsection}{S.\arabic{section}.\arabic{subsection}.\arabic{subsubsection}}

\setcounter{equation}{0}
\renewcommand{\theequation}{s.\arabic{equation}}

\setcounter{figure}{0}
\renewcommand{\thefigure}{S.\arabic{figure}}

\setcounter{table}{0}
\renewcommand{\thetable}{S.\arabic{table}}

\section{Derivation of the Self-Evaluation Loss}
\label{sec:s1-self-eval}

\paragraph{Setup.}
We follow the forward noising in Eq.~(1) and the model
parameterization in Eqs.~(6)–(7) (main paper). Throughout,
we use the same network head $G_\theta$ as in the main text,
and \emph{stop-gradient} is denoted by $\operatorname{sg}[\cdot]$.
All gradients are taken w.r.t.\ $\mathbf{x}_s$ that is obtained by
re-noising the model prediction $\hat{\mathbf{x}}_0$ as in Sec.~3.2.

\paragraph{Posterior means.}
By Tweedie’s formula applied to Eq.~(1), the (data) conditional
and unconditional posterior means, and the (model) conditional
posterior mean, satisfy
\begin{equation}
\label{eq:s-tweedie-cond}
\begin{aligned}
\mathbb{E}_{q}\!\left[\mathbf{x}_0 | \mathbf{x}_s,\mathbf{c}\right]
&= \frac{1}{\alpha_s}\!\left(
\mathbf{x}_s + \sigma_s^2 \nabla_{\mathbf{x}_s}\log q(\mathbf{x}_s|\mathbf{c})
\right),\\
\mathbb{E}_{q}\!\left[\mathbf{x}_0 | \mathbf{x}_s\right]
&= \frac{1}{\alpha_s}\!\left(
\mathbf{x}_s + \sigma_s^2 \nabla_{\mathbf{x}_s}\log q(\mathbf{x}_s)
\right),\\
\mathbb{E}_{p_\theta}\!\left[\mathbf{x}_0 | \mathbf{x}_s,\mathbf{c}\right]
&= \frac{1}{\alpha_s}\!\left(
\mathbf{x}_s + \sigma_s^2 \nabla_{\mathbf{x}_s}\log p_\theta(\mathbf{x}_s|\mathbf{c})
\right).
\end{aligned}
\end{equation}

Subtracting the first two lines of \eqref{eq:s-tweedie-cond} gives
\begin{equation}
\label{eq:s-mean-diff-q}
\begin{aligned}
&\mathbb{E}_{q}\!\left[\mathbf{x}_0 | \mathbf{x}_s\right]
-
\mathbb{E}_{q}\!\left[\mathbf{x}_0 | \mathbf{x}_s,\mathbf{c}\right]
\\
&=
\frac{\sigma_s^2}{\alpha_s}\!\left(
\nabla_{\mathbf{x}_s}\log q(\mathbf{x}_s)
-
\nabla_{\mathbf{x}_s}\log q(\mathbf{x}_s|\mathbf{c})
\right),
\end{aligned}
\end{equation}
and subtracting the first and third lines yields
\begin{equation}
\label{eq:s-mean-diff-model}
\begin{aligned}
&\mathbb{E}_{p_\theta}\!\left[\mathbf{x}_0 | \mathbf{x}_s,\mathbf{c}\right]
-
\mathbb{E}_{q}\!\left[\mathbf{x}_0 | \mathbf{x}_s,\mathbf{c}\right]
\\
&=
\frac{\sigma_s^2}{\alpha_s}\!\left(
\nabla_{\mathbf{x}_s}\log p_\theta(\mathbf{x}_s|\mathbf{c})
-
\nabla_{\mathbf{x}_s}\log q(\mathbf{x}_s|\mathbf{c})
\right).
\end{aligned}
\end{equation}

\subsection{Self-evaluation without auxiliary term}
\label{sec:self-eval-no-aux}

We use the self-evaluation pseudo-target from the main paper,
Eq.~(14),
\[
\mathbf{x}_{\text{self}}
:= \operatorname{sg}\!\left[
\hat{\mathbf{x}}_0 - \bigl(G_\theta(\hat{\mathbf{x}}_s,s,s,\phi)
- G_\theta(\hat{\mathbf{x}}_s,s,s,\mathbf{c})\bigr)
\right],
\]
and the per-sample squared loss (whose expectation over
$(t,s,\mathbf{x}_0,\boldsymbol{\varepsilon})$ gives Eq.~(15)):
\begin{equation}
\label{eq:s-per-sample-selfloss}
\mathcal{L}_{\text{self}} := \left\|
\hat{\mathbf{x}}_0 - \mathbf{x}_{\text{self}}
\right\|_2^{\,2}.
\end{equation}

\paragraph{Result 1.}
Under the posterior-mean approximation
$G_\theta(\mathbf{x}_s,s,s,\mathbf{c})\!\approx\!\mathbb{E}_q[\mathbf{x}_0|\mathbf{x}_s,\mathbf{c}]$ and
$G_\theta(\mathbf{x}_s,s,s,\phi)\!\approx\!\mathbb{E}_q[\mathbf{x}_0|\mathbf{x}_s]$,
the gradient of \eqref{eq:s-per-sample-selfloss} w.r.t.\ $\hat{\mathbf{x}}_s$ is
\begin{equation}
\label{eq:s-grad-basic}
\begin{aligned}
\nabla_{\hat{\mathbf{x}}_s}\,\mathcal{L}_{\text{self}}
&= \Bigl(\tfrac{\partial \hat{\mathbf{x}}_0}{\partial \hat{\mathbf{x}}_s}\Bigr)^{\!\top}
\!\nabla_{\hat{\mathbf{x}}_0}\,\mathcal{L}_{\text{self}}
= \frac{2}{\alpha_s}\!\left(\hat{\mathbf{x}}_0 - \mathbf{x}_{\text{self}}\right)\\
&= \frac{2}{\alpha_s}\!\left(
G_\theta(\hat{\mathbf{x}}_s,s,s,\phi) - G_\theta(\hat{\mathbf{x}}_s,s,s,\mathbf{c})
\right)\\
&\approx \frac{2}{\alpha_s}\!\left(
\mathbb{E}_{q}[\mathbf{x}_0 | \hat{\mathbf{x}}_s]
-
\mathbb{E}_{q}[\mathbf{x}_0 | \hat{\mathbf{x}}_s,\mathbf{c}]
\right)\\
&= \frac{2\sigma_s^2}{\alpha_s^2}\!\left(
\nabla_{\hat{\mathbf{x}}_s}\log q(\hat{\mathbf{x}}_s)
-
\nabla_{\hat{\mathbf{x}}_s}\log q(\hat{\mathbf{x}}_s|\mathbf{c})
\right),
\end{aligned}
\end{equation}
where the last equality uses \eqref{eq:s-mean-diff-q}. Hence gradient
descent on Eq.~(15) moves $\hat{\mathbf{x}}_s$ in the direction of the
\emph{classifier score} $\nabla_{\hat{\mathbf{x}}_s}\log q(\mathbf{c}|\hat{\mathbf{x}}_s)$.

\subsection{Self-evaluation with auxiliary term}
\label{sec:self-eval-aux}

We optionally add a branch prompted by $\mathbf{c}_{\text{fake}}$ to estimate
the model posterior mean
$G_\theta(\mathbf{x}_s,s,s,\mathbf{c}_{\text{fake}})
\approx \mathbb{E}_{p_\theta}[\mathbf{x}_0|\mathbf{x}_s,\mathbf{c}]$.
Define
\begin{equation}
\label{eq:s-Delta}
\begin{aligned}
\Delta_\theta(\mathbf{x}_s,\mathbf{c})
&:=
k\!\left(
G_\theta(\mathbf{x}_s,s,s,\phi)
-
G_\theta(\mathbf{x}_s,s,s,\mathbf{c})
\right)\\
&\quad+
(1-k)\left(
G_\theta(\mathbf{x}_s,s,s,\mathbf{c}_{\text{fake}})
-
G_\theta(\mathbf{x}_s,s,s,\mathbf{c})
\right),
\end{aligned}
\end{equation}
and the target
$
\mathbf{x}_{\text{self}}
:= \operatorname{sg}\!\left[\hat{\mathbf{x}}_0 - \Delta_\theta(\hat{\mathbf{x}}_s,\mathbf{c})\right]$, and the per-sample squared loss $\mathcal{L}_{\text{self}} := \left\|\hat{\mathbf{x}}_0 - \mathbf{x}_{\text{self}}\right\|_2^{\,2}.
$

\paragraph{Result 2.}
Proceeding as in \eqref{eq:s-grad-basic},
\begin{equation}
\label{eq:s-grad-aux}
\begin{aligned}
\nabla_{\hat{\mathbf{x}}_s}\,\mathcal{L}_{\text{self}}
&= \frac{2}{\alpha_s}\,\Delta_\theta(\hat{\mathbf{x}}_s,\mathbf{c})\\
&\hspace{-2.0em}\approx \frac{2}{\alpha_s}\!\left[
k\!\left(\mathbb{E}_{q}[\mathbf{x}_0|\hat{\mathbf{x}}_s]
-\mathbb{E}_{q}[\mathbf{x}_0|\hat{\mathbf{x}}_s,\mathbf{c}]\right)\right.\\
&\hspace{0em}\left.+\;
(1-k)\left(\mathbb{E}_{p_\theta}[\mathbf{x}_0|\hat{\mathbf{x}}_s,\mathbf{c}]
-\mathbb{E}_{q}[\mathbf{x}_0|\hat{\mathbf{x}}_s,\mathbf{c}]
\right)\right]\\
&\hspace{-2.0em}= \frac{2\sigma_s^2}{\alpha_s^2}\!\left[
k\!\left(
\nabla_{\hat{\mathbf{x}}_s}\log q(\hat{\mathbf{x}}_s|\phi)
-
\nabla_{\hat{\mathbf{x}}_s}\log q(\hat{\mathbf{x}}_s|\mathbf{c})
\right)\right.\\
&\hspace{0.0em}\left.+\;
(1-k)\left(\nabla_{\hat{\mathbf{x}}_s}\log p_\theta(\hat{\mathbf{x}}_s|\mathbf{c})
-
\nabla_{\hat{\mathbf{x}}_s}\log q(\hat{\mathbf{x}}_s|\mathbf{c})
\right)\right],
\end{aligned}
\end{equation}
where we used \eqref{eq:s-mean-diff-q} and \eqref{eq:s-mean-diff-model}. Equation
\eqref{eq:s-grad-aux} is proportional to the full ideal vector field
in Eq.~(13) once we set $k=(w-1)/w$. In practice, we set $k=0.9$.

\paragraph{Training.} To realize $G_\theta(\mathbf{x}_s,s,s,\mathbf{c}_{\text{fake}})\!\approx\!
\mathbb{E}_{p_\theta}[\mathbf{x}_0|\mathbf{x}_s,\mathbf{c}]$, we use model
samples and reuse the same conditional FM loss as Eq.~(7): draw
$\hat{\mathbf{x}}_0\!\sim\!p_\theta(\mathbf{x}_0|\mathbf{c})$ and
$\hat{\mathbf{x}}_s\!\sim\!p_\theta(\mathbf{x}_s|\mathbf{x}_0,\mathbf{c})$, and $\mathbf{c}_{\text{fake}}$ is constructed by concatenating the phrase ‘fake image’ with the original prompt, and then minimize
\begin{equation}
\label{eq:s-fake-fm}
\mathcal{L}_{\text{fake}}
=
\mathbb{E}
\!\left[
\left\|G_\theta(\operatorname{sg}[\hat{\mathbf{x}}_s],s,s,\mathbf{c}_{\text{fake}}) - \operatorname{sg}[\hat{\mathbf{x}}_0]\right\|_2^{\,2}
\right].
\end{equation}
In practice we follow the training schedule in Sec.~4 of the main paper:
use only the classifier term early, and enable the auxiliary term later
to refine artifacts, while keeping the overall objective identical to
Eqs.~(16)–(21).
\section{Implementation Details}
\label{sec:s2-implementation}
\begin{figure*}[ht]
  \centering
  \includegraphics[width=0.95\textwidth]{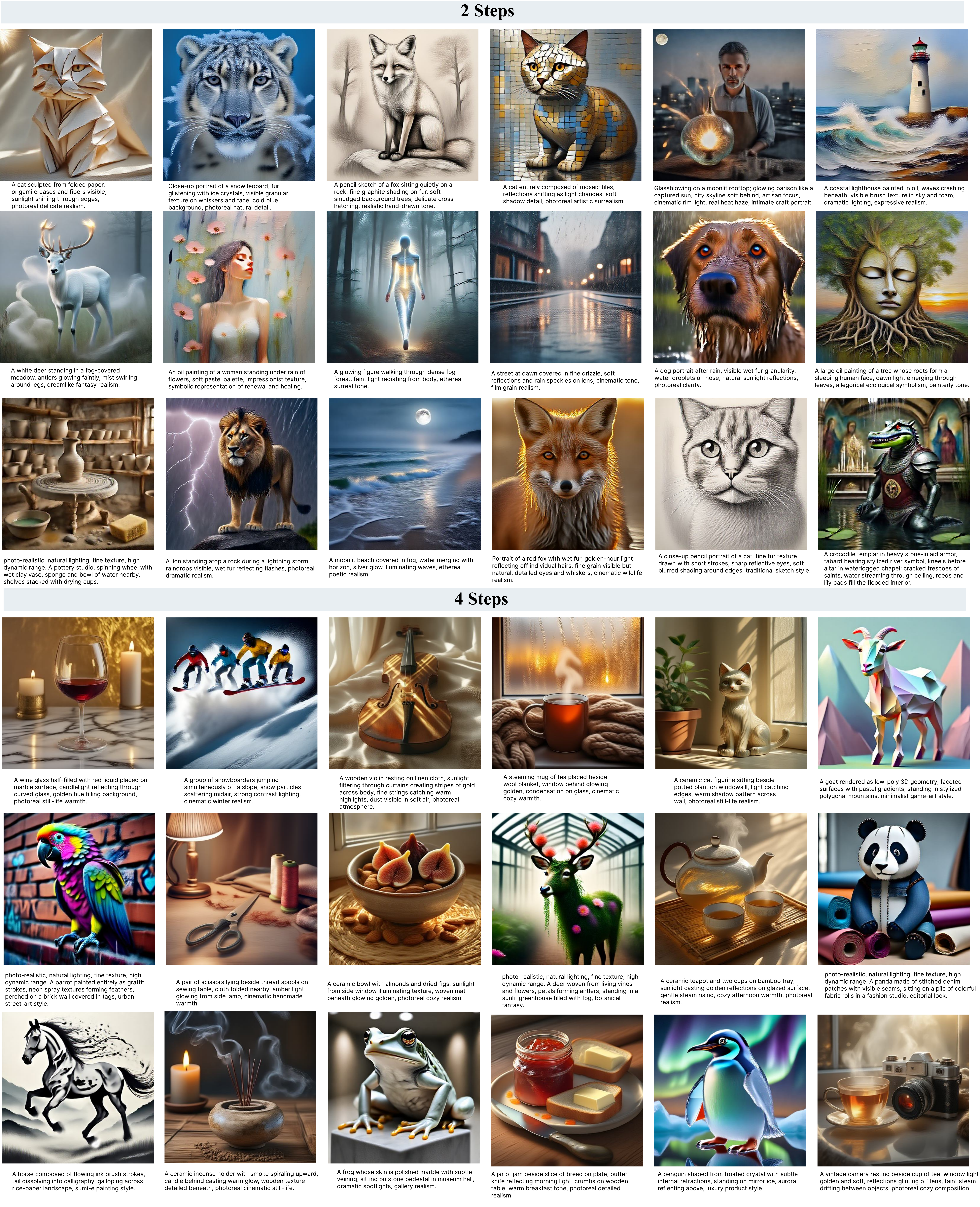}
  \vspace{-0.1cm}
  \caption{\textbf{More results with 2 and 4 steps.} We showcase diverse text-to-image results from our model at 2 and 4 inference step counts,
demonstrating coherent semantics, strong text alignment.}
  \label{fig:2_4_step}
\end{figure*}
\begin{figure*}[ht]
  \centering
  \includegraphics[width=0.95\textwidth]{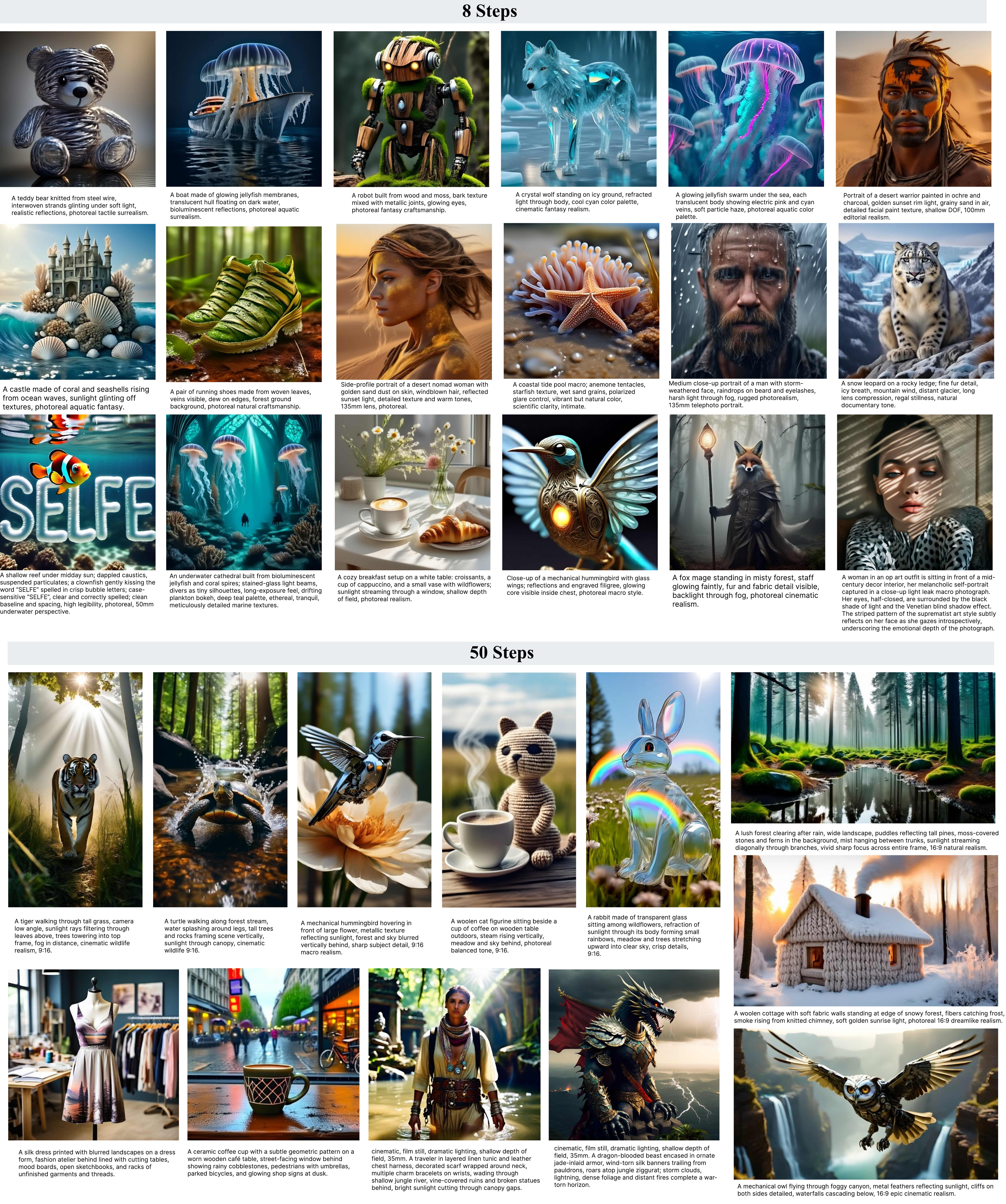}
  \caption{\textbf{More results with 8 and 50 steps.} We showcase diverse text-to-image results from our model at 8 and 50 inference step counts,
demonstrating coherent semantics, strong text alignment. }
  \label{fig:8_50_step}
\end{figure*}

We adopt a latent transformer architecture similar to FLUX~\cite{flux, esser2024scaling} for our experiments, with minor modifications to accommodate our new $s$-input. Specifically, the design of the modules handling $s$ mirrors those handling $t$.

We employ a 2B-parameter model trained on mixed-resolution and varying aspect-ratio text-to-image datasets. Initially, the model is trained at an approximate resolution of $256^2$ pixels for 500k iterations with a batch size of 1024. Subsequently, we introduce higher-resolution data of $512^2$ pixels, maintaining a balanced batch proportion (1:1) between the lower-resolution and higher-resolution data, with a total batch size of 768, continuing training until reaching 710k iterations. At iteration 550k, we additionally introduce training with the auxiliary term. We use the Adam optimizer with $\beta_1=0.9, \beta_2=0.95$, a learning rate warmup for 1000 iterations, and linearly decay the learning rate from $3\times10^{-4}$ to $1\times10^{-5}$. For model evaluation, we maintain an exponential moving average (EMA) with a decay rate of 0.9999. Additionally, during training in the self-evaluation forward pass, the conditional branch utilizes the EMA model, while the unconditional branch employs the non-EMA model.

\paragraph{Architecture.}
We adopt a FLUX-style latent transformer and keep the notation consistent with the main paper: the denoiser's raw prediction is $V_\theta(\cdot)$ and the sample head is
$G_\theta(\mathbf{x}_t,t,s,\mathbf{c})=\mathbf{x}_t - t\,V_\theta(\mathbf{x}_t,t,s,\mathbf{c})$ (cf.\ Eq.~(6)–(7) in the main paper).
Our implementation consists of four modules:
(a) a VAE, (b) a patchifier, (c) frozen text encoders, and (d) a dual-time denoiser.

\textit{(a-b) VAE and patch tokens.}
We use the FLUX.1-dev auto-encoder with $z$-channels $=16$, and compression factors $[1,8,8]$ for $[\text{frames},\text{H},\text{W}]$.
Images are tokenized by a patchifier with patch size $[1,2,2]$.
Thus, each image produces a sequence of $L_\text{img}=(H/16)\times(W/16)$ tokens, each of dimension $d_\text{img}=16\times2\times2=64$.

\textit{(c) Text and global conditioning.}
We use a frozen T5-XXL encoder to obtain token embeddings of dimension $d_\text{txt}=4096$.
Additionally, we compute a global pooled CLIP embedding (ViT-L/14) of dimension $d_\text{vec}=768$.
Both encoders are kept frozen during training, and their outputs are linearly projected to $\mathbb{R}^{d_{\text{model}}}$ before entering the denoiser.

\textit{(d) Denoiser.}
Our 2B model has a
model width $d_\text{model}=2048$, head size $d_\text{head}=128$ (thus $16$ heads),
and a total of $8$ \emph{Double-Stream} blocks followed by $16$ \emph{Single-Stream} blocks.
Positional encoding uses multi-axis RoPE over $(t,y,x)$ with axis dimensions $[16,56,56]$, whose sum matches $d_\text{head}$ and whose three axes correspond to time and the two spatial directions.
Inputs are linearly projected to $d_\text{model}$:
text via $\mathbb{R}^{4096}\!\to\!\mathbb{R}^{2048}$, image tokens via $\mathbb{R}^{64}\!\to\!\mathbb{R}^{2048}$,
and the global CLIP vector via a two-layer MLP $\mathbb{R}^{768}\!\to\!\mathbb{R}^{2048}$.
For ablations, we also train a smaller 0.5B variant with $d_\text{model}=1024$, $d_\text{head}=64$, and RoPE axis dimensions $[8,28,28]$, while keeping all other components identical.

Particularly, the denoiser has two time inputs: the primary time $t$ and an auxiliary time $s$ used by the self-evaluation mechanism (Sec.~3.2). In practice, we encode $t$ and the gap $t-s$ with sinusoidal features followed by small MLPs:
\begin{equation}
\label{eq:s2-time-emb}
e_t = \mathrm{MLP}_t(\mathrm{Sinusoid}(t)),
e_s = \mathrm{MLP}_s(\mathrm{Sinusoid}(t-s)),
\end{equation}
and form a combined time embedding
\begin{equation}
\label{eq:s2-time-emb-combined}
\tilde{e}_t = e_t + e_s.
\end{equation}
This combined embedding $\tilde{e}_t$ simply replaces the original single-time embedding in the backbone: every module that previously consumed $e_t$ now receives $\tilde{e}_t$. Consequently, the only architectural change relative to FLUX is the additional auxiliary term $e_s$ added on top of $e_t$, while all downstream conditioning and modulation remain unchanged.

\paragraph{Timestep Scheduler.}
We first sample the primary time $t$ from a logit-normal distribution defined on $(0,1)$:
\begin{equation}
t_{\text{raw}} = \sigma(z), \quad z \sim \mathcal{N}(0,1),
\end{equation}
where $\sigma(\cdot)$ denotes the sigmoid function. This raw time is further adjusted by a length-dependent warping function. Specifically, given the latent patch length $L$, we define a linear shift $\mu(L)$ interpolating between $0.5$ at length $512$ and $1.15$ at length $4096$, then compute the warped primary time as:
\begin{equation}
\label{eq:warp}
t = \frac{e^{\mu(L)}}{e^{\mu(L)} + (1/t_{\text{raw}} - 1)}.
\end{equation}

For the secondary time $s$, we set $s = t$ with probability $p = 0.5$. For the remaining half of the cases, we sample $s$ uniformly from the interval:
\begin{equation}
s \sim \mathcal{U}((1 - \tau)\,t,\, t),
\end{equation}
where $\tau$ is a linear annealing weight, transitioning from $0$ to $1$ over the first $300,000$ training iterations. As a result, the effective lower bound $(1 - \tau)t$ decreases gradually from approximately $t$ towards $0$ during training. For the weighting function $w_{s,t}$ in Eq.~(20), we set it to $1/t^2$.  

\paragraph{Inference.}
For inference, we employ an initially linear timestep scheduler with a length-dependent warping function, same with \cref{eq:warp}. We use a DDIM-style update with an $\eta$-controlled noise level, following Song et al.~\cite{song2020denoising}; setting $\eta=0$ recovers deterministic DDIM, while $\eta=1$ corresponds to the original DDPM ancestral sampling. In our case, we use $\eta=1$. 
\section{Additional Experimental Results}
\label{sec:s3-experiments}

\subsection{Alternative s-Scheduler}
\label{sec:s3-any-step}
We investigate alternative strategies for selecting the secondary timestep $s_k$ during inference, given a transition from $t_k$ to $t_{k+1}$. During training, the selection of $s_k$ affects two aspects simultaneously: it determines the noise level for the smoothed data distribution used in the reverse KL divergence, and it specifies the self-evaluation weighting factor $\lambda_{s,t}$. These dual roles suggest alternative choices for $s_k$ might yield intermediate and potentially improved behaviors. An intriguing direction for future work would be decoupling the dependence between $s_k$ and the weighting factor $\lambda_{s,t}$, making $\lambda_{s,t}$ independently tunable.

\begin{figure}[t]
  \centering
  \includegraphics[width=0.97\linewidth]{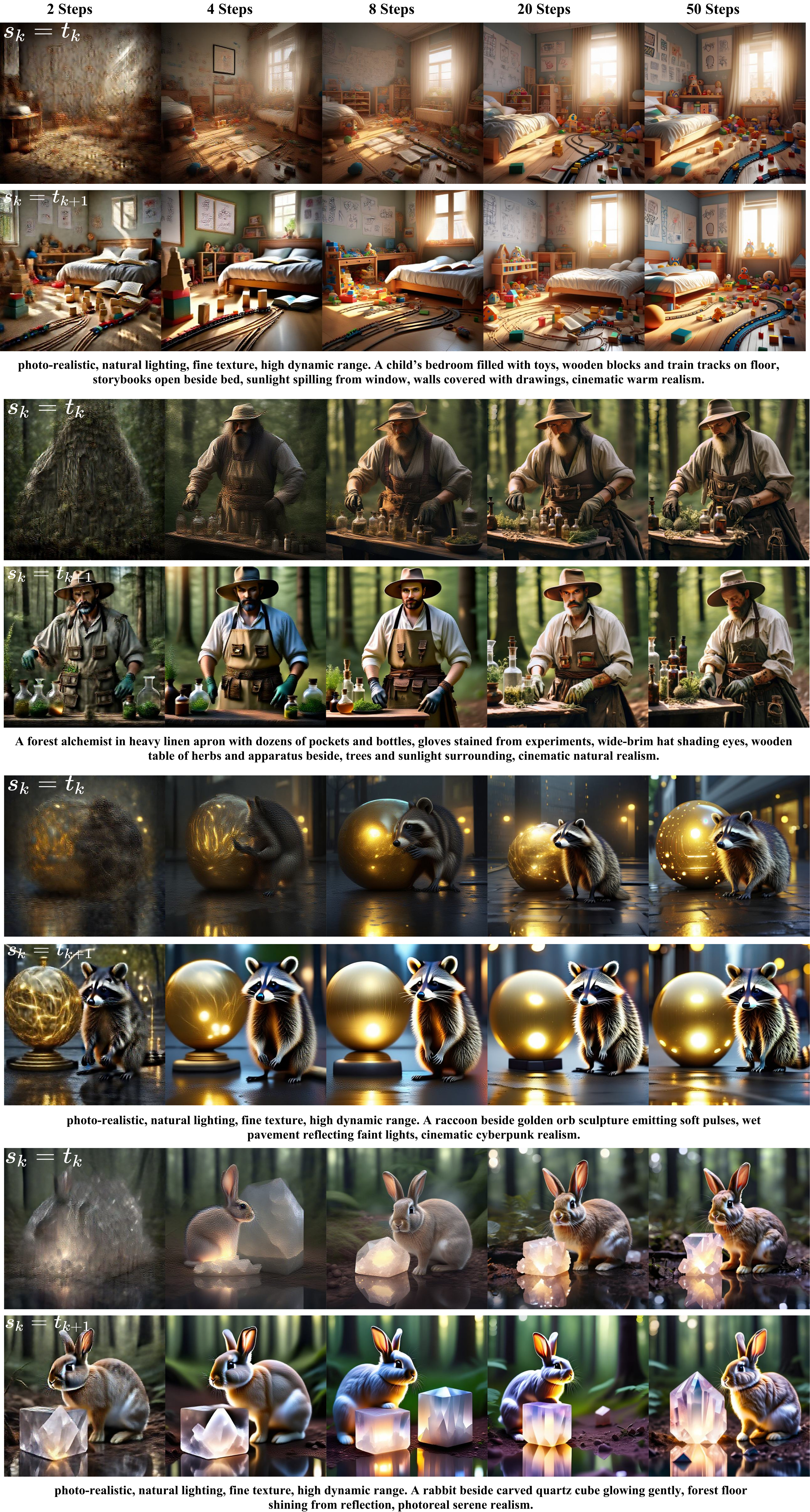}
  \caption{\textbf{Visualization of two special cases for choosing the secondary timestep \( s_k \) during inference.} Top rows: \( s_k = t_k \), bottom rows: \( s_k = t_{k+1} \).} 
  \label{fig:two case}
\end{figure}

We illustrate our empirical observations in \cref{fig:two case}, highlighting two notable special cases:
\begin{enumerate}
    \item When $s_k = t_k$, the model utilizes only the flow matching loss. Consequently, its behavior closely resembles standard Flow Matching, performing poorly at very low inference steps but improving significantly with more steps.
    \item When $s_k = t_{k+1}$, the model excels in few-step generation. However, as the number of inference steps increases (e.g., at 50 steps), we occasionally observe it underperforms compared to $s_k=t_k$ (see the last two examples in \cref{fig:two case}).
\end{enumerate}

\begin{figure}[t]
  \centering
  \includegraphics[width=0.97\linewidth]{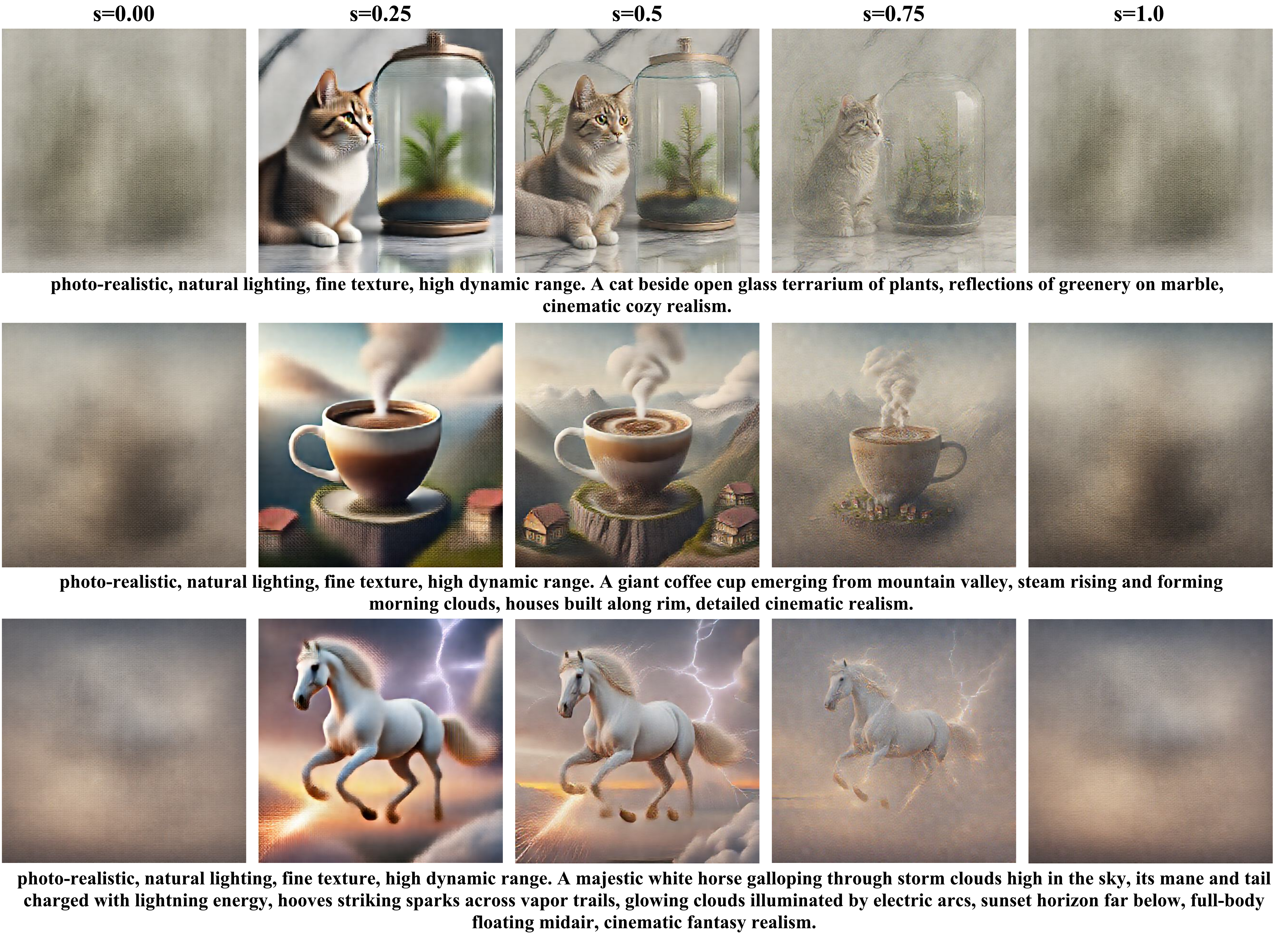}
  \caption{\textbf{One-step generation without classifier-free guidance}. We show results of when selecting different $s$.
  }
  \label{fig:one step}
\end{figure}

Additionally, we explore a special inference setting—\textit{one-step} generation without classifier-free guidance. As shown in \cref{fig:one step}, we interpolate between $s_k = t_k$ (represented as $s=1$) and $s_k = t_{k+1}$ (represented as $s=0$). Both extreme cases fail to yield meaningful images, whereas the midpoint choice $s=0.5$ achieves a favorable balance between texture detail and overall image coherence.


\subsection{More Results}

We present more results at different inference budgets in \cref{fig:2_4_step} and \cref{fig:8_50_step}.
\section{Prompts of Results}
\label{sec:s4-prompt}

We provide the text prompts used for the qualitative results shown in the main paper.

\subsection{Prompts of Figure 1.}

\noindent\textbf{2-step:}
\begin{itemize}
    \item The word ``Self-E'' appearing faintly through condensation on a train window, blurred landscape passing behind, city lights refracting, cinematic melancholic tone.
    \item Portrait of a wolf under snowfall, frost collecting on its muzzle and fur, visible texture and natural grain, calm expression, photoreal cold-environment realism.
    \item An oil painting of a woman with her hair turning into waves, seascape blending with portrait, tactile brushwork, painterly surreal tone.
\end{itemize}

\noindent\textbf{4-step:}
\begin{itemize}
    \item A volcano erupting with petals instead of lava, clouds of color drifting across the sky, surreal cinematic beauty.
    \item A cat composed of smoke sitting on a rooftop, its form dissolving into the night air, glowing eyes reflecting city lights, detailed cinematic surrealism.
    \item A plate of pastries beside a teacup, sunlight highlighting golden crusts, powdered sugar shimmering under a warm glow, photoreal comforting realism.
\end{itemize}

\noindent\textbf{8-step:}
\begin{itemize}
    \item Portrait of a jungle guardian with vine tattoos and green-gold war paint, wet skin glistening under filtered sunlight, 85mm, macro detail on skin texture, cinematic naturalism.
    \item A bison standing in a foggy grassland at dawn, dew on tall grass, sun barely visible through haze, fur glistening with moisture, cinematic atmospheric realism.
    \item A cozy cottage built entirely from red and white yarn, knitted walls and woven roof shingles, soft texture visible in each thread, golden sunlight casting gentle shadows, photoreal tactile realism.
\end{itemize}

\noindent\textbf{50-step:}
\begin{itemize}
    \item A human face emerging from cracked porcelain, half side smooth and half crumbled revealing crystalline interior, emotional surreal realism.
    \item A queen in jeweled crown standing under golden archway, sunlight refracting through gems, detailed embroidery on gown, distant cityscape visible behind, regal photoreal tone, 9:16.
    \item A rabbit made of transparent glass jumping across a shallow creek, sunlight refracting rainbow light through its body, ripples and stones visible beneath, forest on both sides, 16:9 photoreal wide scene.
    \item A close-up underwater portrait of a woman leaning forward on a large rectangular glowing sign that reads “Self-E,” the sign filling the lower part of the frame like a real physical board. Neon hues of cyan, pink, and gold from the illuminated surface ripple through the clear turquoise water, casting colorful reflections across her face. She smiles brightly, blue eyes open with confidence, freckles and natural skin texture visible under shifting light. Transparent fish swim nearby among coral branches, tiny bubbles rising through the calm cinematic 9:16 scene.
    \item A valley full of blooming lupines and daisies, 16:9 panoramic view, rolling hills leading toward mountain horizon, warm afternoon light highlighting color contrast, photoreal cinematic realism.
\end{itemize}

\subsection{Prompts of Figure 4.}
\begin{itemize}
    \item A colorful chalkboard artwork spelling “SELFE” in bright pastel colors—blue, pink, yellow, and green—each letter outlined softly, chalk dust particles floating through air, faint eraser marks around, warm nostalgic classroom atmosphere.
    \item A small home bar setup with wine bottles, glass of whiskey half full, sliced lemon on napkin, reflections on wooden counter, photoreal cinematic tone.
    \item A cat sleeping on cloud drifting above mountain range, soft pink sunrise illuminating fur, photoreal dreamlike realism.
    \item A royal guard in ornate jade armor, sword reflecting sunlight, palace gardens behind full of flowers and fountains, silk banners waving in soft breeze, cinematic elegant realism.
\end{itemize}

\subsection{Prompts of Figure 5.}
\begin{itemize}
    \item A high-altitude thunderhead above a wheat plain; sculpted cumulonimbus, sunlit anvil, tiny barn for scale, global contrast, 24mm vastness, dramatic meteorological realism.
    \item A house constructed from luminous jelly bricks glowing at night, detailed transparency and refraction, cinematic realism.
\end{itemize}

\section{Limitations and Future Work}
\label{sec:s5-limitation}

While our method significantly surpasses existing from-scratch training methods in few-step generation, it still has some limitations. Notably, our current approach, although effective in significantly reducing the number of inference steps, cannot fully compete with the quality obtained by 50-step inference when employing extremely few steps (e.g., 1–2 steps). In these cases, the generated images may lack sufficiently sharp details.

Additionally, given that our proposed paradigm fundamentally differs from existing consistency-based methods, it remains at an early stage of exploration. Several critical design choices, such as loss weighting schemes and inference strategies, have not yet been thoroughly optimized. We believe further systematic exploration of these aspects could lead to considerable improvements.

Nonetheless, we emphasize that our method introduces a genuinely novel training paradigm, distinct from the consistency-training family. Empirically, we observe that our method inherently produces robust structure and semantic coherence, exhibiting a clear trend of generating coherent structures first, followed by iterative refinement of details.

Looking forward, we identify several promising avenues for future work:
\begin{enumerate}
\item Improving training strategies and inference-time scheduling to further enhance generation quality.
\item Investigating the efficacy of our approach for downstream task fine-tuning.
\item Exploring scalability and potential adaptations of the proposed paradigm to video generative models.
\item Extending our method to unconditional generative settings, as the current approach relies on conditional guidance to derive the classifier scores.
\end{enumerate}

\end{document}